\documentclass[]{interact}

\usepackage{epstopdf}
\usepackage{subfigure}

\usepackage{natbib}
\bibpunct[, ]{(}{)}{;}{a}{}{,}

\theoremstyle{plain}

\theoremstyle{definition}

\theoremstyle{remark}

\usepackage{lscape}
\usepackage{hyperref}
\usepackage[utf8]{inputenc}

\usepackage{setspace}
\usepackage{booktabs}
\usepackage{longtable}
\usepackage{array}
\usepackage{multirow}
\usepackage{wrapfig}
\usepackage{float}
\usepackage{colortbl}
\usepackage{pdflscape}
\usepackage{tabu}
\usepackage{threeparttable}
\usepackage{threeparttablex}
\usepackage[normalem]{ulem}
\usepackage{makecell}
\usepackage{xcolor}

\begin{document}

\articletype{}

\title{Automated Assessment of Residual Plots with Computer Vision
Models}

\author{\name{Weihao Li$^{a}$, Dianne Cook$^{a}$, Emi Tanaka$^{a, b,
c}$, Susan VanderPlas$^{d}$, Klaus Ackermann$^{a}$}
\affil{$^{a}$Department of Econometrics and Business Statistics, Monash
University, Clayton, VIC, Australia; $^{b}$Biological Data Science
Institute, Australian National University, Acton, ACT,
Australia; $^{c}$Research School of Finance, Actuarial Studies and
Statistics, Australian National University, Acton, ACT,
Australia; $^{d}$Department of Statistics, University of Nebraska,
Lincoln, Nebraska, USA}
}

\thanks{CONTACT Weihao
Li. Email: \href{mailto:weihao.li@monash.edu}{\nolinkurl{weihao.li@monash.edu}}, Dianne
Cook. Email: \href{mailto:dicook@monash.edu}{\nolinkurl{dicook@monash.edu}}, Emi
Tanaka. Email: \href{mailto:emi.tanaka@anu.edu.au}{\nolinkurl{emi.tanaka@anu.edu.au}}, Susan
VanderPlas. Email: \href{mailto:susan.vanderplas@unl.edu}{\nolinkurl{susan.vanderplas@unl.edu}}, Klaus
Ackermann. Email: \href{mailto:klaus.ackermann@monash.edu}{\nolinkurl{klaus.ackermann@monash.edu}}}

\maketitle

\begin{abstract}
Plotting the residuals is a recommended procedure to diagnose deviations
from linear model assumptions, such as non-linearity,
heteroscedasticity, and non-normality. The presence of structure in
residual plots can be tested using the lineup protocol to do visual
inference. There are a variety of conventional residual tests, but the
lineup protocol, used as a statistical test, performs better for
diagnostic purposes because it is less sensitive and applies more
broadly to different types of departures. However, the lineup protocol
relies on human judgment which limits its scalability. This work
presents a solution by providing a computer vision model to automate the
assessment of residual plots. It is trained to predict a distance
measure that quantifies the disparity between the residual distribution
of a fitted classical normal linear regression model and the reference
distribution, based on Kullback-Leibler divergence. From extensive
simulation studies, the computer vision model exhibits lower sensitivity
than conventional tests but higher sensitivity than human visual tests.
It is slightly less effective on non-linearity patterns. Several
examples from classical papers and contemporary data illustrate the new
procedures, highlighting its usefulness in automating the diagnostic
process and supplementing existing methods.
\end{abstract}


\section{Introduction}\label{sec-model-introduction}

Plotting residuals is commonly regarded as a standard practice in linear
regression diagnostics \citep{belsley1980regression, cook1982residuals}.
This visual assessment plays a crucial role in identifying whether model
assumptions, such as linearity, homoscedasticity, and normality, are
reasonable. It also helps in understanding the goodness of fit and
various unexpected characteristics of the model.

\begin{figure}[!h]

{\centering \includegraphics[width=1\linewidth]{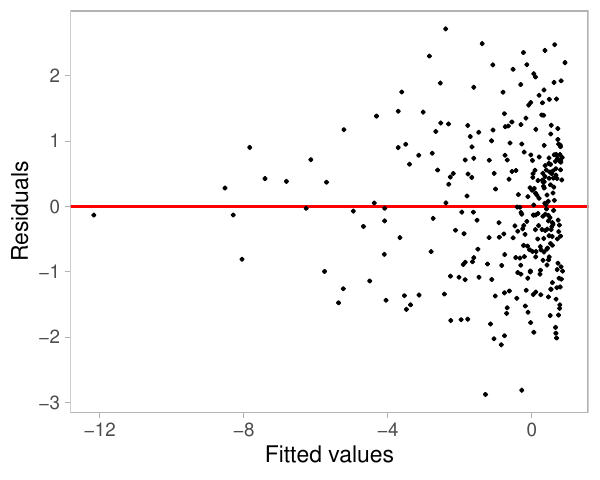} 

}

\caption{An example residual vs fitted values plot (red line indicates 0 corresponds to the x-intercept, i.e. $y=0$). The vertical spread of the data points varies with the fitted values. This often indicates the existence of heteroskedasticity, however, here the result is due to skewed distribution of the predictors rather than heteroskedasticity. The Breusch-Pagan test rejects this residual plot at 95\% significance level ($p\text{-value} = 0.046$).}\label{fig:false-finding}
\end{figure}

Generating a residual plot in most statistical software is often as
straightforward as executing a line of code or clicking a button.
However, accurately interpreting a residual plot can be challenging. A
residual plot can exhibit various visual features, but it is crucial to
recognize that some may arise from the characteristics of predictors and
the natural stochastic variation of the observational unit, rather than
indicating a violation of model assumptions \citep{li2024plot}. Consider
Figure \ref{fig:false-finding} as an example, the residual plot displays
a triangular left-pointing shape. The distinct difference in the spread
of the residuals across the fitted values may result in the analyst
suggesting that there may be heteroskedasticity, however, it is
important to avoid over-interpreting this visual pattern. In this case,
the fitted regression model is correctly specified, and the triangular
shape is actually a result of the skewed distribution of the predictors,
rather than indicating a flaw in the model.

The concept of visual inference, as proposed by
\citet{buja2009statistical}, provides an inferential framework to assess
whether residual plots indeed contain visual patterns inconsistent with
the model assumptions. The fundamental idea involves testing whether the
true residual plot visually differs significantly from null plots, where
null plots are plotted with residuals generated from the residual
rotation distribution \citep{langsrud2005rotation}, which is a
distribution consistent with the null hypothesis \(H_0\) that the linear
regression model is correctly specified. Typically, the visual test is
accomplished through the lineup protocol, where the true residual plot
is embedded within a lineup alongside several null plots. If the true
residual plot can be distinguished from the lineup, it provides evidence
for rejecting \(H_0\).

The practice of delivering a residual plot as a lineup is generally
regarded as a valuable approach. Beyond its application in residual
diagnostics, the lineup protocol has been integrated into the analysis
of diverse subjects. For instance, Loy and Hofmann
\citetext{\citeyear{loy2013diagnostic}; \citeyear{loy2014hlmdiag}; \citeyear{loy2015you}}
illustrated its applicability in diagnosing hierarchical linear models.
Additionally, \citet{widen2016graphical} and \citet{fieberg2024using}
demonstrated its utility in geographical and ecology research
respectively, while \citet{krishnan2021hierarchical} explored its
effectiveness in forensic examinations.

A practical limitation of the lineup protocol lies in its reliance on
human judgements \citep[see][ about the practical
limitations]{li2024plot}. Unlike conventional statistical tests that can
be performed computationally in statistical software, the lineup
protocol requires human evaluation of images. This characteristic makes
it less suitable for large-scale applications, given the associated high
labour costs and time requirements. There is a substantial need to
develop an approach to substitute these human judgement with an
automated reading of data plots using machines.

The utilization of computers to interpret data plots has a rich history,
with early efforts such as ``Scagnostics'' by \citet{tukey1985computer},
a set of numerical statistics that summarize features of scatter plots.
\citet{wilkinson2005graph} expanded on this work, introducing
scagnostics based on computable measures applied to planar proximity
graphs. These measures, including, but not limited to, ``Outlying'',
``Skinny'', ``Stringy'', ``Straight'', ``Monotonic'', ``Skewed'',
``Clumpy'', and ``Striated'', aimed to characterize outliers, shape,
density, trend, coherence and other characteristics of the data. While
this approach has been inspiring, there is a recognition
\citep{buja2009statistical} that it may not capture all the necessary
visual features that differentiate true residual plots from null plots.
A more promising alternative entails enabling machines to learn the
function for extracting visual features from residual plots.
Essentially, this means empowering computers to discern the crucial
visual features for residual diagnostics and determining the method to
extract them.

Modern computer vision models are well-suited for addressing this
challenge. They rely on deep neural networks with convolutional layers
\citep{fukushima1982neocognitron}. These layers use small, sliding
windows to scan the image, performing a dot product to extract local
features and patterns. Numerous studies have demonstrated the efficacy
of convolutional layers in addressing various vision tasks, including
image recognition \citep{rawat2017deep}. Despite the widespread use of
computer vision models in fields like computer-aided diagnosis
\citep{lee2015image}, pedestrian detection \citep{brunetti2018computer},
and facial recognition \citep{emami2012facial}, their application in
reading data plots remains limited. While some studies have explored the
use of computer vision models for tasks such as reading recurrence plots
for time series regression \citep{ojeda2020multivariate}, time series
classification
\citep{chu2019automatic, hailesilassie2019financial, hatami2018classification, zhang2020encoding},
anomaly detection \citep{chen2020convolutional}, and pairwise causality
analysis \citep{singh2017deep}, the application of reading residual
plots with computer vision models is a new field of study.

In this paper, we develop computer vision models and integrate them into
the residual plots diagnostics workflow, addressing the need for an
automated visual inference. The paper is structured as follows. Section
\ref{sec-model-specifications} discusses various specifications of the
computer vision models. Section
\ref{sec-model-distance-between-residual-plots} defines the distance
measure used to detect model violations, while Section
\ref{sec-model-distance-estimation} explains how the computer vision
models estimate this distance measure. Section
\ref{sec-model-statistical-testing} covers the statistical tests based
on the estimated distance, and Section \ref{sec-model-violations-index}
introduces a Model Violations Index, which offers a quicker and more
convenient assessment. Sections \ref{sec-model-data-generation},
\ref{sec-model-architecture}, and \ref{sec-model-training} detail the
data preparation, model architecture, and training process,
respectively. The results are presented in Section
\ref{sec-model-results}. Example dataset applications are discussed in
Section \ref{sec-examples}. Finally, we conclude with a discussion of
our findings and propose ideas for future research directions.

\section{Model Specifications}\label{sec-model-specifications}

There are various specifications of the computer vision model that can
be used to assess residual plots. We discuss these specifications below
focusing on two key components of the model formula: the input and the
output format.

\subsection{Input Formats}\label{input-formats}

Deep learning models are in general very sensitive to the input data.
The quality and relevance of the input data greatly influence the
model's capacity to generate insightful and meaningful results. There
are several designs of the input format that can be considered.

A straightforward architecture of the input layer involves feeding a
vector of residuals along with a vector of fitted values, essentially
providing all the necessary information for creating a residuals vs
fitted values plot. However, a drawback of this method is the dynamic
input size, which changes based on the number of observations. For
modern computer vision models implemented in mainstream software like
TensorFlow \citep{abadi2016tensorflow}, the input shape is typically
fixed. One solution is to pad the input vectors with leading or trailing
zeros when the input tensor expects longer vectors, but it may fail if
the input vector surpasses the designed length.

Another strategy is to summarize the residuals and fitted values
separately using histograms and utilize the counts as the input. By
controlling the number of bins in the histograms, it becomes possible to
provide fixed-length input vectors. Still, since histograms only capture
the marginal distribution of residuals and fitted values respectively,
they can not be used to differentiate visual patterns with same marginal
distributions but different joint distributions.

Another architecture of the input layer involves using an image as
input. The primary advantage of this design, as opposed to the vector
format, is the availability of the existing and sophisticated image
processing architectures developed over the years, such as the VGG16
architecture proposed in \citet{simonyan2014very}. These architectures
can effectively capture and summarize spatial information from nearby
pixels, which is less straightforward with vector input. While encoding
data points as pixels might lead to some loss of detail, this approach
still provides a more comprehensive representation than the
histogram-based strategy because it captures the joint distribution of
fitted values and residuals. The main considerations are the image
resolution and the aesthetics of the residual plot. In general, a higher
resolution provides more information to the model but comes with a
trade-off of increased complexity and greater difficulty in training. As
for the aesthetics of the residual plot, a practical solution is to
consistently present residual plots in the same style to the model. This
implies that the model can not accept arbitrary images as input but
requires the use of the same pre-processing pipeline to convert
residuals and fitted values into a standardized-style residual plot.

Providing multiple residual plots to the model, such as a pair of plots,
a triplet or a lineup is also a possible option.
\citet{chopra2005learning} have shown that computer vision models
designed for image comparison can assess whether a pair of images are
similar or dissimilar. Applied to our specific problem, we can define
null plots of a fitted regression model to be similar to each other,
while considering true residual plots to be distinct from null plots of
any fitted regression model. A triplet constitutes a set of three
images, denoted as \(image_1\), \(image_2\) and \(image_3\). It is often
used to predict whether \(image_2\) or \(image_3\) is more similar to
\(image_1\), proving particularly useful for establishing rankings
between samples. For this setup, we can apply the same criteria to
define similarity between images. However, it is important to note that
these two approaches usually require additional considerations regarding
the loss function and, at times, non-standard training processes due to
shared weights between different convolutional blocks.

Presenting a lineup to a model aligns closely with the lineup protocol.
However, as the number of residual plots in a lineup increases, the
resolution of the input image grows rapidly, posing challenges in
training the model. We experimented with this approach in a pilot study,
but the performance of the trained model was sub-optimal.

Taking into account the implementation cost and the need for model
interpretability, we used the single residual plot input format in this
paper.

\subsection{Output Formats}\label{output-formats}

Given that the input is a single residual plot represented as a
fixed-resolution image, we can choose the output from the computer
vision model to be either binary (classification) or numeric
(regression).

The binary outcome can represent whether the input image is consistent
with a null plot as determined by either (1) the data generating process
or (2) the result of a visual test based on human judgement. Training a
model following the latter option requires data from prior human subject
experiments, presenting difficulties in controlling the quality of data
due to variations in experimental settings across different studies.
Additionally, some visual inference experiments are unrelated to linear
regression models or residual plot diagnostics, resulting in a limited
amount of available training data.

Alternatively, the output could be a meaningful and interpretable
numerical measure useful for assessing residual plots, such as the
strength of suspicious visual patterns reflecting the extent of model
violations, or the difficulty index for identifying whether a residual
plot has no issues. However, these numeric measures are often informally
used in daily communication but are not typically formalized or
rigorously defined. For the purpose of training a model, this numeric
measure has to be quantifiable.

In this study, we chose to define and use a distance between a true
residual plot and a theoretically ``good'' residual plot. This is
further explained in Section
\ref{sec-model-distance-between-residual-plots}.
\citet{vo2016localizing} have also demonstrated that defining a proper
distance between images can enhance the matching accuracy in image
search compared to a binary outcome model.

\subsection{Auxiliary Information with
Scagnostics}\label{auxiliary-information-with-scagnostics}

In Section \ref{sec-model-introduction}, we mention that scagnostics
consist of a set of manually designed visual feature extraction
functions. While our computer vision model will learn its own feature
extraction function during training, leveraging additional information
from scagnostics can enhance the model's predictive accuracy.

For each residual plot used as an input image, we calculated four
scagnostics --- ``Monotonic'', ``Sparse'', ``Splines'', and ``Striped''
-- using the \texttt{cassowaryr} R package \citep{mason2022cassowaryr}.
These computed measures, along with the number of observations from the
fitted model, were provided as the second input for the computer vision
model. We selected these scagnostics due to their reliability and
efficiency, as other scagnostics occasionally caused R process crashes
(approximately 5\% of the time) during training data preparation, due to
a bug in the \texttt{interp} R package \citep{Albrecht2023interp}.
Although the package maintainer later fixed this bug at our request, the
fix came too late to retrain the model, and additionally, their high
computational costs make them unsuitable for rapid inference, which was
a critical factor in our choice.

\section{Distance from a Theoretically ``Good'' Residual
Plot}\label{sec-model-distance-between-residual-plots}

To develop a computer vision model for assessing residual plots within
the visual inference framework, it is important to precisely define a
numerical measure of ``difference'' or ``distance'' between plots. This
distance can take the form of a basic statistical operation on pixels,
such as the sum of square differences, however, a pixel-to-pixel
comparison makes little sense in comparing residual plots where the main
interest would be structural patterns. Alternatively, it could involve
established image similarity metrics like the Structural Similarity
Index Measure \citep{wang2004image} which compares images by integrating
three perception features of an image: contrast, luminance, and
structure (related to average, standard deviation and correlation of
pixel values over a window, respectively). These image similarity
metrics are tailored for image comparison in vastly different tasks to
evaluating data plots, where only essential plot elements require
assessment \citep{chowdhury2018measuring}. We can alternatively define a
notion of distance by integrating key plot elements (instead of key
perception features like luminance, contrast, and structure), such as
those captured by scagnostics mentioned in Section
\ref{sec-model-introduction}, but the functional form still needs to be
carefully refined to accurately reflect the extent of the violations.

In this section, we introduce a distance measure between a true residual
plot and a theoretically `good' residual plot. This measure quantifies
the divergence between the residual distribution of a given fitted
regression model and that of a correctly specified model. The
computation assumes knowledge of the data generating processes for
predictors and response variables. Since these processes are often
unknown in practice, we will discuss a method to estimate this distance
using a computer vision model in Section
\ref{sec-model-distance-estimation}.

\subsection{Residual Distribution}\label{residual-distribution}

For a classical normal linear regression model,
\(\boldsymbol{y} = \boldsymbol{X}\boldsymbol{\beta} + \boldsymbol{e}\),
the residual \(\hat{\boldsymbol{e}}\) are derived as the difference of
the fitted values and observed values \(\boldsymbol{y}\). Suppose the
data generating process is known and the regression model is correctly
specified, by the Frisch-Waugh-Lowell theorem \citep{frisch1933partial},
residuals \(\hat{\boldsymbol{e}}\) can also be treated as random
variables and written as a linear transformation of the error
\(\boldsymbol{e}\) formulated as
\(\hat{\boldsymbol{e}} = \boldsymbol{R}\boldsymbol{e}\), where
\(\boldsymbol{R}=\boldsymbol{I}_n -\boldsymbol{X}(\boldsymbol{X}^\top\boldsymbol{X})^{-1}\boldsymbol{X}^\top\)
is the residual operator, \(\boldsymbol{I}_n\) is a \(n\) by \(n\)
identity matrix, and \(n\) is the number of observations.

One of the assumptions of the classical normal linear regression model
is that the error \(\boldsymbol{e}\) follows a multivariate normal
distribution with zero mean and constant variance, i.e.,
\(\boldsymbol{e} \sim N(\boldsymbol{0}_n,\sigma^2\boldsymbol{I}_n)\). It
follows that the distribution of residuals \(\hat{\boldsymbol{e}}\) can
be characterized by a certain probability distribution, denoted as
\(Q\), which is transformed from the multivariate normal distribution.
This reference distribution \(Q\) summarizes what ``good'' residuals
should follow given the design matrix \(\boldsymbol{X}\) is known and
fixed.

Suppose the design matrix \(\boldsymbol{X}\) has linearly independent
columns, the trace of the hat matrix
\(\boldsymbol{H} = \boldsymbol{X}(\boldsymbol{X}^\top\boldsymbol{X})^{-1}\boldsymbol{X}^\top\)
will equal to the number of columns in \(\boldsymbol{X}\) denoted as
\(k\). As a result, the rank of \(\boldsymbol{R}\) is \(n - k\), and
\(Q\) is a degenerate multivariate distribution. To capture the
characteristics of \(Q\), such as moments, we can simulate a large
numbers of \(\boldsymbol{\varepsilon}\) and transform it to
\(\boldsymbol{e}\) to get the empirical estimates. For simplicity, in
this study, we replaced the variance-covariance matrix of residuals
\(\text{cov}(\boldsymbol{e}, \boldsymbol{e}) = \boldsymbol{R}\sigma^2\boldsymbol{R}^\top = \boldsymbol{R}\sigma^2\)
with a full-rank diagonal matrix
\(\text{diag}(\boldsymbol{R}\sigma^2)\), where \(\text{diag}(.)\) sets
the non-diagonal entries of a matrix to zeros. The resulting
distribution for \(Q\) is
\(N(\boldsymbol{0}_n, \text{diag}(\boldsymbol{R}\sigma^2))\).

Distribution \(Q\) is derived from the correctly specified model.
However, if the model is misspecified, then the actual distribution of
residuals denoted as \(P\), will be different from \(Q\). For example,
if the data generating process contains variables correlated with any
column of \(\boldsymbol{X}\) but missing from \(\boldsymbol{X}\),
causing an omitted variable problem, \(P\) will be different from \(Q\)
because the residual operator obtained from the fitted regression model
will not be the same as \(\boldsymbol{R}\). Besides, if the
\(\boldsymbol{\varepsilon}\) follows a non-normal distribution such as a
multivariate lognormal distribution, \(P\) will usually be skewed and
has a long tail.

\subsection{\texorpdfstring{Distance of \(P\) from
\(Q\)}{Distance of P from Q}}\label{distance-of-p-from-q}

Defining a proper distance between distributions is usually easier than
defining a proper distance between data plots. Given the true residual
distribution \(Q\) and the reference residual distribution \(P\), we
used a distance measure based on Kullback-Leibler divergence
\citep{kullback1951information} to quantify the difference between two
distributions as

\begin{equation} \label{eq:kl-0}
D = \log\left(1 + D_{KL}\right),
\end{equation}

where \(D_{KL}\) is defined as

\begin{equation} \label{eq:kl-1}
D_{KL} = \int_{\mathbb{R}^{n}}\log\frac{p(\boldsymbol{e})}{q(\boldsymbol{e})}p(\boldsymbol{e})d\boldsymbol{e},
\end{equation}

\noindent and \(p(.)\) and \(q(.)\) are the probability density
functions for distribution \(P\) and distribution \(Q\), respectively.

This distance measure was first proposed in \citet{li2024plot}. It was
mainly designed for measuring the effect size of non-linearity and
heteroskedasticity in a residual plot. \citet{li2024plot} have derived
that, for a classical normal linear regression model that omits
necessary higher-order predictors \(\boldsymbol{Z}\) and the
corresponding parameter \(\boldsymbol{\beta}_z\), and incorrectly
assumes
\(\boldsymbol{\varepsilon} \sim N(\boldsymbol{0}_n,\sigma^2\boldsymbol{I}_n)\)
while in fact
\(\boldsymbol{\varepsilon} \sim N(\boldsymbol{0}_n, \boldsymbol{V})\)
where \(\boldsymbol{V}\) is an arbitrary symmetric positive
semi-definite matrix, \(Q\) can be represented as
\(N(\boldsymbol{R}\boldsymbol{Z}\boldsymbol{\beta}_z, \text{diag}(\boldsymbol{R}\boldsymbol{V}\boldsymbol{R}))\).
Note that the variance-covariance matrix is replaced with the diagonal
matrix to ensure it is a full-rank matrix.

Since both \(P\) and \(Q\) are adjusted to be multivariate normal
distributions, Equation \ref{eq:kl-1} can be further expanded to

\begin{equation} \label{eq:kl-2}
D_{KL} = \frac{1}{2}\left(\log\frac{|\boldsymbol{W}|}{|\text{diag}(\boldsymbol{R}\sigma^2)|} - n + \text{tr}(\boldsymbol{W}^{-1}\text{diag}(\boldsymbol{R}\sigma^2)) + \boldsymbol{\mu}_z^\top\boldsymbol{W}^{-1}\boldsymbol{\mu}_z\right),
\end{equation}

\noindent where
\(\boldsymbol{\mu}_z = \boldsymbol{R}\boldsymbol{Z}\boldsymbol{\beta}_z\),
and
\(\boldsymbol{W} = \text{diag}(\boldsymbol{R}\boldsymbol{V}\boldsymbol{R})\).
The assumed error variance \(\sigma^2\) is set to be
\(\text{tr}(\boldsymbol{V})/n\), which is the expectation of the
estimated variance.

\subsection{\texorpdfstring{Non-normal
\(P\)}{Non-normal P}}\label{non-normal-p}

For non-normal error \(\boldsymbol{\varepsilon}\), the true residual
distribution \(P\) is unlikely to be a multivariate normal distribution.
Thus, Equation \ref{eq:kl-2} given in \citet{li2024plot} will not be
applicable to models violating the normality assumption.

To evaluate the Kullback-Leibler divergence of non-normal \(P\) from
\(Q\), the fallback is to solve Equation \ref{eq:kl-1} numerically.
However, since \(\boldsymbol{e}\) is a linear transformation of
non-normal random variables, it is very common that the general form of
\(P\) is unknown, meaning that we can not easily compute
\(p(\boldsymbol{e})\) using a well-known probability density function.
Additionally, even if \(p(\boldsymbol{e})\) can be calculated for any
\(\boldsymbol{e} \in \mathbb{R}^n\), it will be very difficult to do
numerical integration over the \(n\)-dimensional space, because \(n\)
could be potentially very large.

In order to approximate \(D_{KL}\) in a practically computable manner,
the elements of \(\boldsymbol{e}\) are assumed to be independent of each
other. This assumption solves both of the issues mentioned above. First,
we no longer need to integrate over \(n\) random variables. The result
of Equation \ref{eq:kl-1} is now the sum of the Kullback-Leibler
divergence evaluated for each individual residual due to the assumption
of independence between observations. Second, it is not required to know
the joint probability density \(p(\boldsymbol{e})\) any more. Instead,
the evaluation of Kullback-Leibler divergence for an individual residual
relies on the knowledge of the marginal density \(p_i(e_i)\), where
\(e_i\) is the \(i\)-th residual for \(i = 1, ..., n\). This is much
easier to approximate through simulation. It is also worth mentioning
that this independence assumption generally will not hold if
\(\text{cov}(e_i, e_j) \neq 0\) for any \(1 \leq i < j \leq n\), but its
existence is essential for reducing the computational cost.

Given \(\boldsymbol{X}\) and \(\boldsymbol{\beta}\), the algorithm for
approximating Equation \ref{eq:kl-1} starts from simulating \(m\) sets
of observed values \(\boldsymbol{y}\) according to the data generating
process. The observed values are stored in a matrix \(\boldsymbol{A}\)
with \(n\) rows and \(m\) columns, where each column of
\(\boldsymbol{A}\) is a set of observed values. Then, we can get \(m\)
sets of realized values of \(\boldsymbol{e}\) stored in the matrix
\(\boldsymbol{B}\) by applying the residual operator
\(\boldsymbol{B} = \boldsymbol{R}\boldsymbol{A}\). Furthermore, kernel
density estimation (KDE) with Gaussian kernel and optimal bandwidth
selected by the Silverman's rule of thumb \citep{silverman2018density}
is applied on each row of \(\boldsymbol{B}\) to estimate \(p_i(e_i)\)
for \(i = 1, ..., n\). The KDE computation can be done by the
\texttt{density} function in R.

Since the Kullback-Leibler divergence can be viewed as the expectation
of the log-likelihood ratio between distribution \(P\) and distribution
\(Q\) evaluated on distribution \(P\), we can reuse the simulated
residuals in matrix \(\boldsymbol{B}\) to estimate the expectation by
the sample mean. With the independence assumption, for non-normal \(P\),
\(D_{KL}\) can be approximated by

\begin{align*} \label{eq:kl-3}
D_{KL} &\approx \sum_{i = 1}^{n} \hat{D}_{KL}^{(i)}, \\
\hat{D}_{KL}^{(i)} &= \frac{1}{m}\sum_{j = 1}^{m} \log\frac{\hat{p}_i(B_{ij})}{q(B_{ij})},
\end{align*}

\noindent where \(\hat{D}_{KL}^{(i)}\) is the estimator of the
Kullback-Leibler divergence for an individual residual \(e_i\),
\(\boldsymbol{B}_{ij}\) is the \(i\)-th row and \(j\)-th column entry of
the matrix \(\boldsymbol{B}\), \(\hat{p}_i(.)\) is the kernel density
estimator of \(p_i(.)\), \(q(.)\) is the normal density function with
mean zero and an assumed variance estimated as
\(\hat{\sigma}^2 = \sum_{b \in vec(\boldsymbol{B})}(b - \sum_{b \in vec(\boldsymbol{B})} b/nm)^2/(nm - 1)\),
and \(vec(.)\) is the vectorization operator which turns a
\(n \times m\) matrix into a \(nm \times 1\) column vector by stacking
the columns of the matrix on top of each other.

\section{Distance Estimation}\label{sec-model-distance-estimation}

In the previous sections, we have defined a distance measure given in
Equation \ref{eq:kl-0} for quantifying the difference between the true
residual distribution \(P\) and an ideal reference distribution \(Q\).
However, this distance measure can only be computed when the data
generating process is known. In reality, we often have no knowledge
about the data generating process, otherwise we do not need to do a
residual diagnostic in the first place.

We use a computer vision model to estimate this distance measure for a
residual plot. Let \(D\) be the result of Equation \ref{eq:kl-0}, and
our estimator \(\hat{D}\) is formulated as

\begin{equation} \label{eq:d-approx}
\hat{D} = f_{CV}(V_{h \times w}(\boldsymbol{e}, \hat{\boldsymbol{y}})),
\end{equation}

\noindent where \(V_{h \times w}(.)\) is a plotting function that saves
a residuals vs fitted values plot with fixed aesthetic as an image with
\(h \times w\) pixels in three channels (RGB), \(f_{CV}(.)\) is a
computer vision model which takes an \(h \times w\) image as input and
predicts the distance in the domain \([0, +\infty)\).

With the estimated distance \(\hat{D}\), we can compare the underlying
distribution of the residuals to a theoretically ``good'' residual
distribution. \(\hat{D}\) can also be used as an index of the model
violations indicating the strength of the visual signal embedded in the
residual plot.

It is not expected that \(\hat{D}\) will be equal to original distance
\(D\). This is largely because information contained in a single
residual plot is limited and it may not be able to summarize all the
important characteristics of the residual distribution. For a given
residual distribution \(P\), many different residual plots can be
simulated, where many will share similar visual patterns, but some of
them could be visually very different from the rest, especially for
regression models with small \(n\). This suggests the error of the
estimation will vary depends on whether the input residual plot is
representative or not.

\section{Statistical Testing}\label{sec-model-statistical-testing}

\subsection{Lineup Evaluation}\label{sec-model-lineup-evaluation}

Theoretically, the distance \(D\) for a correctly specified model is
\(0\), because \(P\) will be the same as \(Q\). However, the computer
vision model may not necessary predict \(0\) for a null plot. Using
Figure \ref{fig:false-finding} as an example, it contains a visual
pattern which is an indication of heteroskedasticity. We would not
expect the model to be able to magically tell if the suspicious pattern
is caused by the skewed distribution of the fitted values or the
existence of heteroskedasticity. Additionally, some null plots could
have outliers or strong visual patterns due to randomness, and a
reasonable model will try to summarize those information into the
prediction, resulting in \(\hat{D} > 0\).

This property is not an issue if \(\hat{D} \gg 0\) for which the visual
signal of the residual plot is very strong, and we usually do not need
any further examination of the significance of the result. However, if
the visual pattern is weak or moderate, having \(\hat{D}\) will not be
sufficient to determine if \(H_0\) should be rejected.

To address this issue we can adhere to the paradigm of visual inference,
by comparing the estimated distance \(\hat{D}\) to the estimated
distances for the null plots in a lineup. Specifically, if a lineup
comprises 20 plots, the null hypothesis \(H_0\) will be rejected if
\(\hat{D}\) exceeds the maximum estimated distance among the \(m - 1\)
null plots, denoted as
\(\max\limits_{1 \leq i \leq m-1} {\hat{D}_{null}^{(i)}}\), where
\(\hat{D}_{null}^{(i)}\) represents the estimated distance for the
\(i\)-th null plot. This approach is equivalent to the typical lineup
protocol requiring a 95\% significance level, where \(H_0\) is rejected
if the data plot is identified as the most distinct plot by the sole
observer. The estimated distance serves as a metric to quantify the
difference between the data plot and the null plots, as intended.

Moreover, if the number of plots in a lineup, denoted by \(m\), is
sufficiently large, the empirical distribution of
\({\hat{D}_{null}^{(i)}}\) can be viewed as an approximation of the null
distribution of the estimated distance. Consequently, quantiles of the
null distribution can be estimated using the sample quantiles, and these
quantiles can be utilized for decision-making purposes. The details of
the sample quantile computation can be found in
\citet{hyndman1996sample}. For instance, if \(\hat{D}\) is greater than
or equal to the 95\% sample quantile, denoted as \(Q_{null}(0.95)\), we
can conclude that the estimated distance for the true residual plot is
significantly different from the estimated distance for null plots with
a 95\% significance level. Based on our experience, to obtain a stable
estimate of the 95\% quantile, the number of null plots, \(n_{null}\),
typically needs to be at least 100. However, if the null distribution
exhibits a long tail, a larger number of null plots may be required.
Alternatively, a \(p\)-value is the probability of observing a distance
equally or greater than \(\hat{D}\) under the null hypothesis \(H_0\),
and it can be estimated by
\(\frac{1}{m} + \frac{1}{m}\sum_{i=1}^{m-1}I\left(\hat{D}_{null}^{(i)} \geq \hat{D}\right)\).

To alleviate computation burden, a lattice of quantiles for \(\hat{D}\)
under \(H_0\) with specified sample sizes can be precomputed. We can
then map the \(\hat{D}\) and sample size to the closet quantile and
sample size in lattice to calculate the corresponding \(p\)-value. This
approach lose precision in \(p\)-value calculation, however,
significantly improves computational efficiency.

\subsection{Bootstrapping}\label{bootstrapping}

Bootstrap is often employed in linear regression when conducting
inference for estimated parameters \citep[see][ and
\citet{efron1994introduction}]{davison1997bootstrap}. It is typically
done by sampling individual cases with replacement and refitting the
regression model. If the observed data accurately reflects the true
distribution of the population, the bootstrapped estimates can be used
to measure the variability of the parameter estimate without making
strong distributional assumptions about the data generating process.

Similarly, bootstrap can be applied on the estimated distance
\(\hat{D}\). For each refitted model \(M_{boot}^{(i)}\), there will be
an associated residual plot \(V_{boot}^{(i)}\) which can be fed into the
computer vision model to obtain \(\hat{D}_{boot}^{(i)}\), where
\(i = 1,...,n_{boot}\), and \(n_{boot}\) is the number of bootstrapped
samples. If we are interested in the variation of \(\hat{D}\), we can
use \(\hat{D}_{boot}^{(i)}\) to estimate a confidence interval.

Alternatively, since each \(M_{boot}^{(i)}\) has a set of estimated
coefficients \(\hat{\boldsymbol{\beta}}_{boot}^{(i)}\) and an estimated
variance \(\hat{\sigma^2}_{boot}^{(i)}\), a new approximated null
distribution can be construed and the corresponding 95\% sample quantile
\(Q_{boot}^{(i)}(0.95)\) can be computed. Then, if
\(\hat{D}_{boot}^{(i)} \geq Q_{boot}^{(i)}(0.95)\), \(H_0\) will be
rejected for \(M_{boot}^{(i)}\). The ratio of rejected
\(M_{boot}^{(i)}\) among all the refitted models provides an indication
of how often the assumed regression model are considered to be incorrect
if the data can be obtained repetitively from the same data generating
process. But this approach is computationally very expensive because it
requires \(n_{boot} \times n_{null}\) times of residual plot assessment.
In practice, \(Q_{null}(0.95)\) can be used to replace
\(Q_{boot}^{(i)}(0.95)\) in the computation.

\section{Model Violations Index}\label{sec-model-violations-index}

While statistical testing is a powerful tool for detecting model
violations, it can become cumbersome and time-consuming when quick
decisions are needed, particularly due to the need to evaluate numerous
null plots. In practice, a more convenient and immediate method for
assessing model performance is often required. This is where an index,
such as the Model Violations Index (MVI), becomes valuable. It offers a
straightforward way to quantify deviations from model assumptions,
enabling rapid assessment and easier comparison across models.

The estimator \(\hat{D}\) measures the difference between the true
residual distribution and the reference residual distribution, a
difference primarily arises from deviations in model assumptions. The
magnitude of \(D\) directly reflects the degree of these deviations,
thus making \(\hat{D}\) instrumental in forming a model violations index
(MVI).

Note that if more observations are used for estimating the linear
regression, the result of Equation \ref{eq:kl-1} will increase, as the
integration will be performed with larger \(n\). For a given data
generating process, \(D\) typically increases logarithmically with the
number of observations. This behavior comes from the relationship
\(D = \text{log}(1 + D_{KL})\), where
\(D_{KL} = \sum_{i=1}^{n}D_{KL}^{(i)}\) under the assumption of
independence.

Since \(\hat{D}\) is an estimate of \(D\), it is expected that a larger
number of observations will also lead to a higher \(\hat{D}\). However,
this does not imply that \(\hat{D}\) fails to accurately represent the
extent of model violations. In fact, when examining residual plots with
more observations, we often observe a stronger visual signal strength,
as the underlying patterns are more likely to be revealed, except in
cases of significant overlapping.

Therefore, the Model Violations Index (MVI) can be proposed as

\begin{equation} \label{eq:mvi}
\text{MVI} = C + \hat{D} - \log(n),
\end{equation}

\noindent where \(C\) is a large enough constant keeping the result
positive and the term \(-\log(n)\) is used to offset the increase in
\(D\) due to sample size.

\begin{figure}[!h]

{\centering \includegraphics[width=1\linewidth]{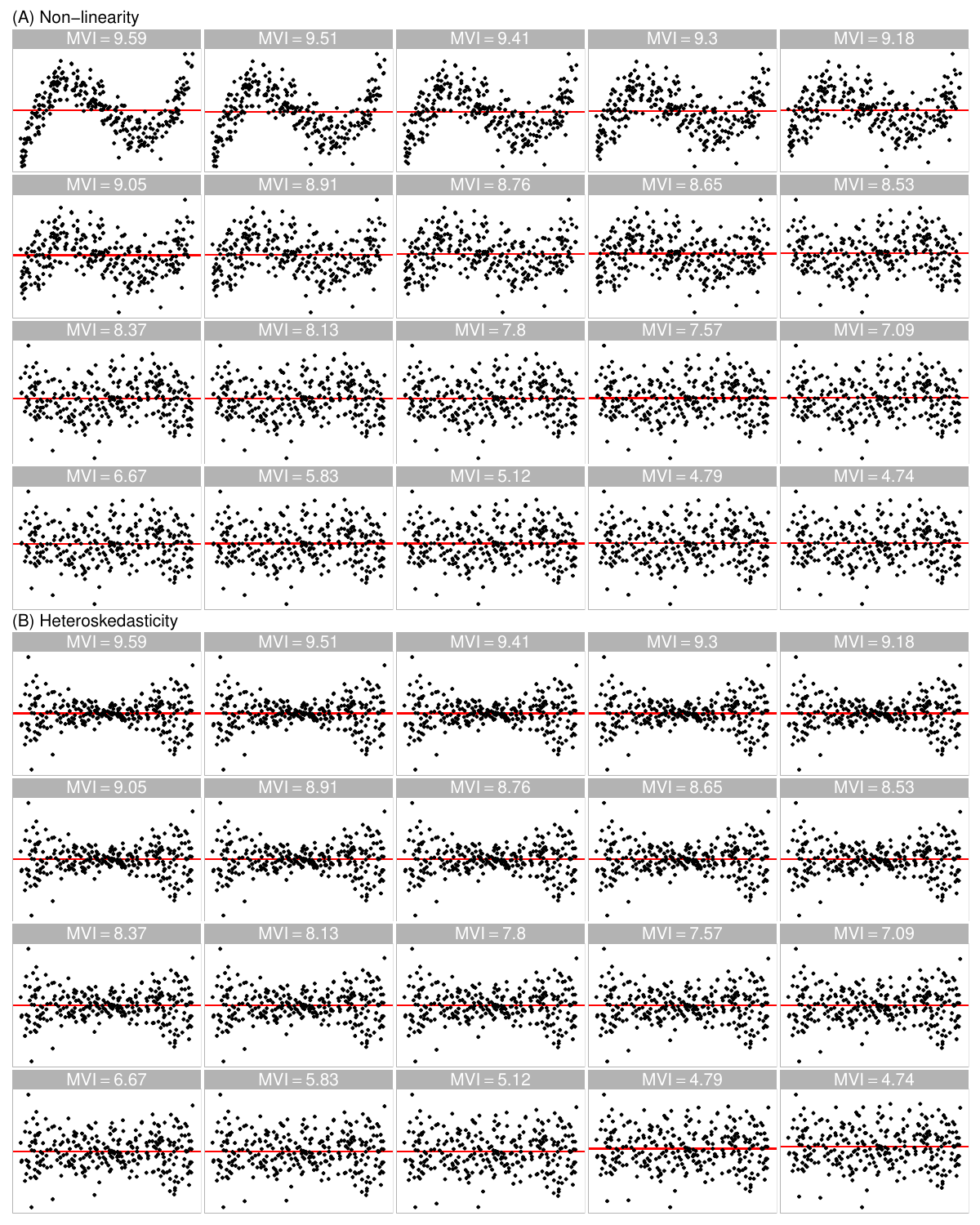} 

}

\caption{Residual plots generated from fitted models exhibiting varying degrees of (A) non-linearity and (B) heteroskedasticity violations. The model violations index (MVI) is displayed atop each residual plot. The non-linearity patterns are relatively strong for $MVI > 8$, and relatively weak for $MVI < 6$, while the heteroskedasticity patterns are relatively strong for $MVI > 8$, and relatively weak for $MVI < 6$.}\label{fig:poly-heter-index}
\end{figure}

Figure \ref{fig:poly-heter-index} displays the residual plots for fitted
models exhibiting varying degrees of non-linearity and
heteroskedasticity. Each residual plot's MVI is computed using Equation
\ref{eq:mvi} with \(C = 10\). When \(\text{MVI} > 8\), the visual
patterns are notably strong and easily discernible by humans. In the
range \(6 < \text{MVI} < 8\), the visibility of the visual pattern
diminishes as MVI decreases. Conversely, when \(\text{MVI} < 6\), the
visual pattern tends to become relatively faint and challenging to
observe. Table \ref{tab:mvi} provides a summary of the MVI usage and it
is applicable to other linear regression models.

\begin{table}

\caption{\label{tab:mvi}Degree of model violations or the strength of the visual signals according to the Model Violations Index (MVI). The constant $C$ is set to be 10.}
\centering
\begin{tabular}[t]{lc}
\toprule
Degree of model violations & Range ($C$ = 10)\\
\midrule
Strong & $\text{MVI} > 8$\\
Moderate & $6 < \text{MVI} < 8$\\
Weak & $\text{MVI} < 6$\\
\bottomrule
\end{tabular}
\end{table}

\section{Data Generation}\label{sec-model-data-generation}

\subsection{Simulation Scheme}\label{simulation-scheme}

While observational data is frequently employed in training models for
real-world applications, the data generating process of observational
data often remains unknown, making computation for our target variable
\(D\) unattainable. Consequently, the computer vision models developed
in this study were trained using synthetic data, including 80,000
training images and 8,000 test images. This approach provided us with
precise label annotations. Additionally, it ensured a large and diverse
training dataset, as we had control over the data generating process,
and the simulation of the training data was relatively cost-effective.

We have incorporated three types of residual departures of linear
regression model in the training data, including non-linearity,
heteroskedasticity and non-normality. All three departures can be
summarized by the data generating process formulated as

\begin{align} \label{eq:data-sim}
\boldsymbol{y} &= \boldsymbol{1}_n + \boldsymbol{x}_1 + \beta_1\boldsymbol{x}_2 + \beta_2(\boldsymbol{z} + \beta_1\boldsymbol{w}) + \boldsymbol{k} \odot \boldsymbol{\varepsilon}, \\
\boldsymbol{z} &= \text{He}_j(g(\boldsymbol{x}_1, 2)), \\
\boldsymbol{w} &= \text{He}_j(g(\boldsymbol{x}_2, 2)), \\
\boldsymbol{k} &= \left[\boldsymbol{1}_n + b(2 - |a|)(\boldsymbol{x}_1 + \beta_1\boldsymbol{x}_2 - a\boldsymbol{1}_n)^{\circ2}\right]^{\circ1/2},
\end{align}

\noindent where \(\boldsymbol{y}\), \(\boldsymbol{x}_1\),
\(\boldsymbol{x}_2\), \(\boldsymbol{z}\), \(\boldsymbol{w}\),
\(\boldsymbol{k}\) and \(\boldsymbol{\varepsilon}\) are vectors of size
\(n\), \(\boldsymbol{1}_n\) is a vector of ones of size \(n\),
\(\boldsymbol{x}_1\) and \(\boldsymbol{x}_2\) are two independent
predictors, \(\text{He}_j(.)\) is the \(j\)th-order probabilist's
Hermite polynomials \citep{hermite1864nouveau}, \((.)^{\circ2}\) and
\((.)^{\circ1/2}\) are Hadamard square and square root, \(\odot\) is the
Hadamard product, and \(g(\boldsymbol{x}, k)\) is a scaling function to
enforce the support of the random vector to be \([-k, k]^n\) defined as

\[g(\boldsymbol{x}, k) = 2k \cdot \frac{\boldsymbol{x} - x_{\min}\boldsymbol{1}_n}{x_{\max} - x_{\min}} - k\boldsymbol{1}_n,~for~k > 0,\]
\noindent where \(x_{\min} = \underset{i \in \{ 1,...,n\}}{\min} x_i\),
\(x_{\max} = \underset{i \in \{ 1,...,n\}}{\max} x_i\) and \(x_i\) is
the \(i\)-th entry of \(\boldsymbol{x}\).

\begin{table}

\caption{\label{tab:factor}Factors used in the data generating process for synthetic data simulation. Factor $j$ and $a$ controls the non-linearity shape and the heteroskedasticity shape respectively. Factor $b$, $\sigma_\varepsilon$ and $n$ control the signal strength. Factor $\text{dist}_\varepsilon$, $\text{dist}_{x1}$ and $\text{dist}_{x2}$ specifies the distribution of $\varepsilon$, $X_1$ and $X_2$ respectively.}
\centering
\begin{tabular}[t]{ll}
\toprule
Factor & Domain\\
\midrule
j & \{2, 3, ..., 18\}\\
a & {}[-1, 1]\\
b & {}[0, 100]\\
$\beta_1$ & \{0, 1\}\\
$\beta_2$ & \{0, 1\}\\
\addlinespace
$\text{dist}_\varepsilon$ & \{discrete, uniform, normal, lognormal\}\\
$\text{dist}_{x1}$ & \{discrete, uniform, normal, lognormal\}\\
$\text{dist}_{x2}$ & \{discrete, uniform, normal, lognormal\}\\
$\sigma_{\varepsilon}$ & {}[0.0625, 9]\\
$\sigma_{X1}$ & {}[0.3, 0.6]\\
\addlinespace
$\sigma_{X2}$ & {}[0.3, 0.6]\\
n & {}[50, 500]\\
\bottomrule
\end{tabular}
\end{table}

The residuals and fitted values of the fitted model were obtained by
regressing \(\boldsymbol{y}\) on \(\boldsymbol{x}_1\). If
\(\beta_1 \neq 0\), \(\boldsymbol{x}_2\) was also included in the design
matrix. This data generation process was adapted from
\citet{li2024plot}, where it was utilized to simulate residual plots
exhibiting non-linearity and heteroskedasticity visual patterns for
human subject experiments. A summary of the factors utilized in Equation
\ref{eq:data-sim} is provided in Table \ref{tab:factor}.

\begin{figure}[!h]

{\centering \includegraphics[width=1\linewidth]{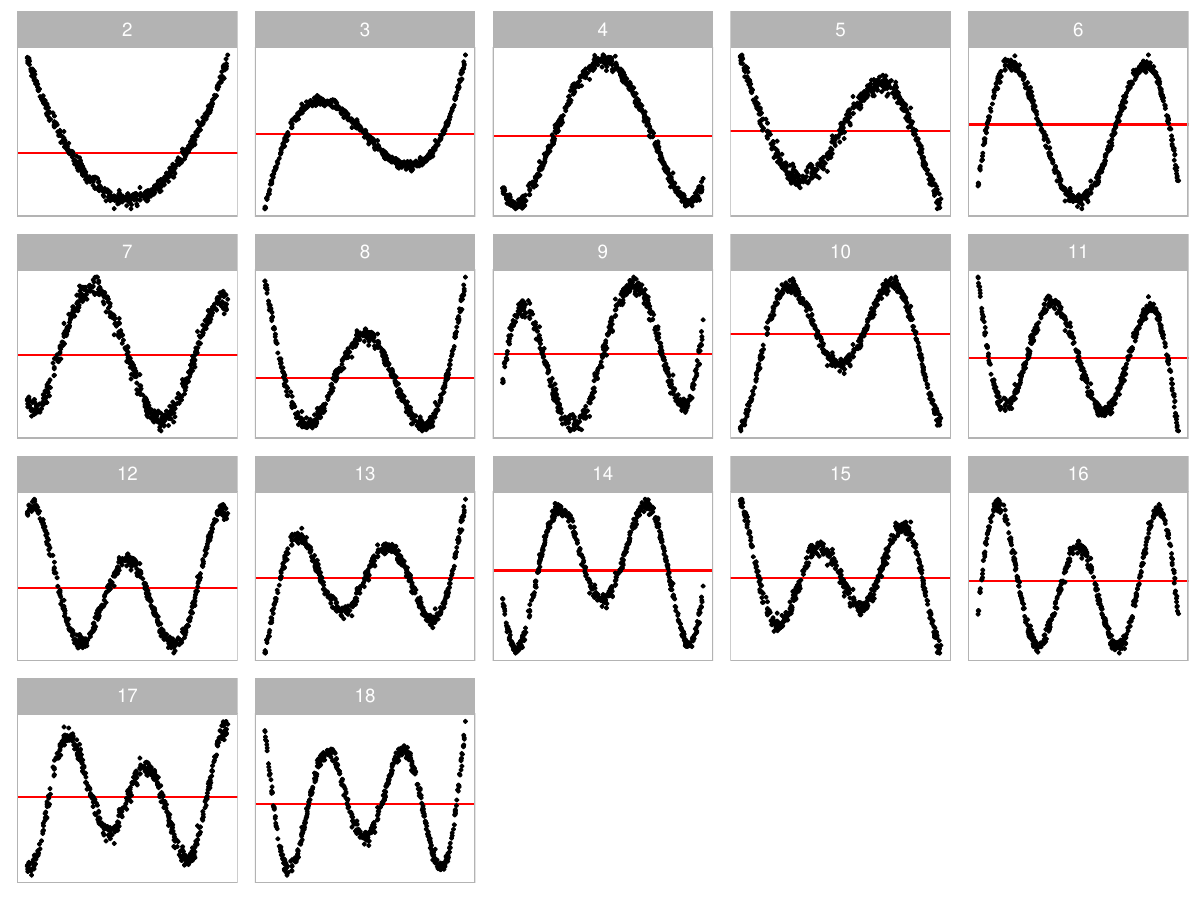} 

}

\caption{Non-linearity forms generated for the synthetic data simulation. The 17 shapes are generated by varying the order of polynomial given by $j$ in $He_j(.)$.}\label{fig:different-j}
\end{figure}

In Equation \ref{eq:data-sim}, \(\boldsymbol{z}\) and \(\boldsymbol{w}\)
represent higher-order terms of \(\boldsymbol{x}_1\) and
\(\boldsymbol{x}_2\), respectively. If \(\beta_2 \neq 0\), the
regression model will encounter non-linearity issues. Parameter \(j\)
serves as a shape parameter that controls the number of tuning points in
the non-linear pattern. Typically, higher values of \(j\) lead to an
increase in the number of tuning points, as illustrated in Figure
\ref{fig:different-j}.

\begin{figure}[!h]

{\centering \includegraphics[width=1\linewidth]{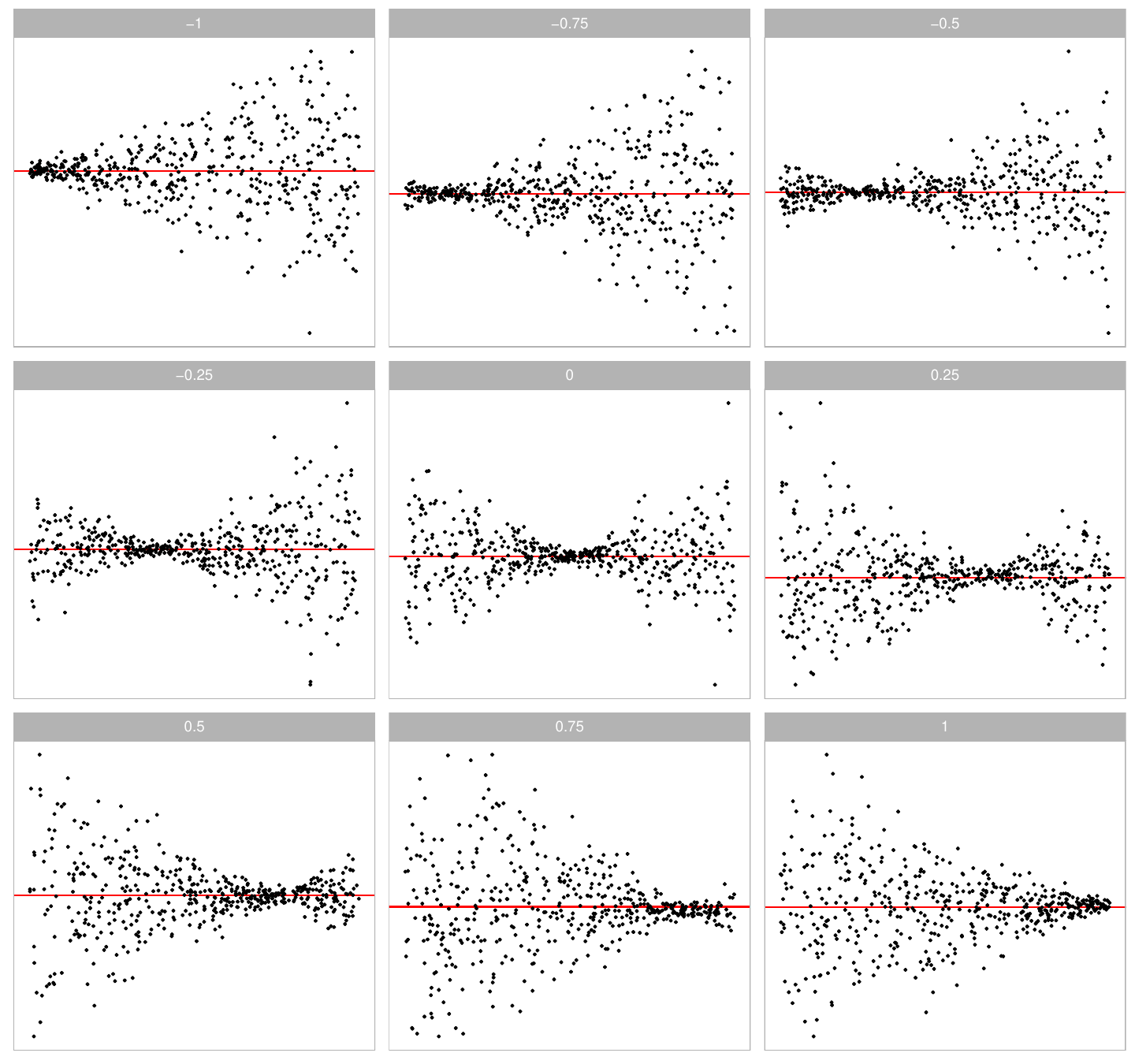} 

}

\caption{Heteroskedasticity forms generated for the synthetic data simulation. Different shapes are controlled by the continuous factor $a$ between -1 and 1. For $a = -1$, the residual plot exhibits a "left-triangle" shape. And for $a = 1$, the residual plot exhibits a "right-triangle" shape. }\label{fig:different-a}
\end{figure}

Additionally, scaling factor \(\boldsymbol{k}\) directly affects the
error distribution and it is correlated with \(\boldsymbol{x}_1\) and
\(\boldsymbol{x}_2\). If \(b \neq 0\) and
\(\boldsymbol{\varepsilon} \sim N(\boldsymbol{0}_n, \sigma^2\boldsymbol{I}_n)\),
the constant variance assumption will be violated. Parameter \(a\) is a
shape parameter controlling the location of the smallest variance in a
residual plot as shown in Figure \ref{fig:different-a}.

\begin{figure}[!h]

{\centering \includegraphics[width=1\linewidth]{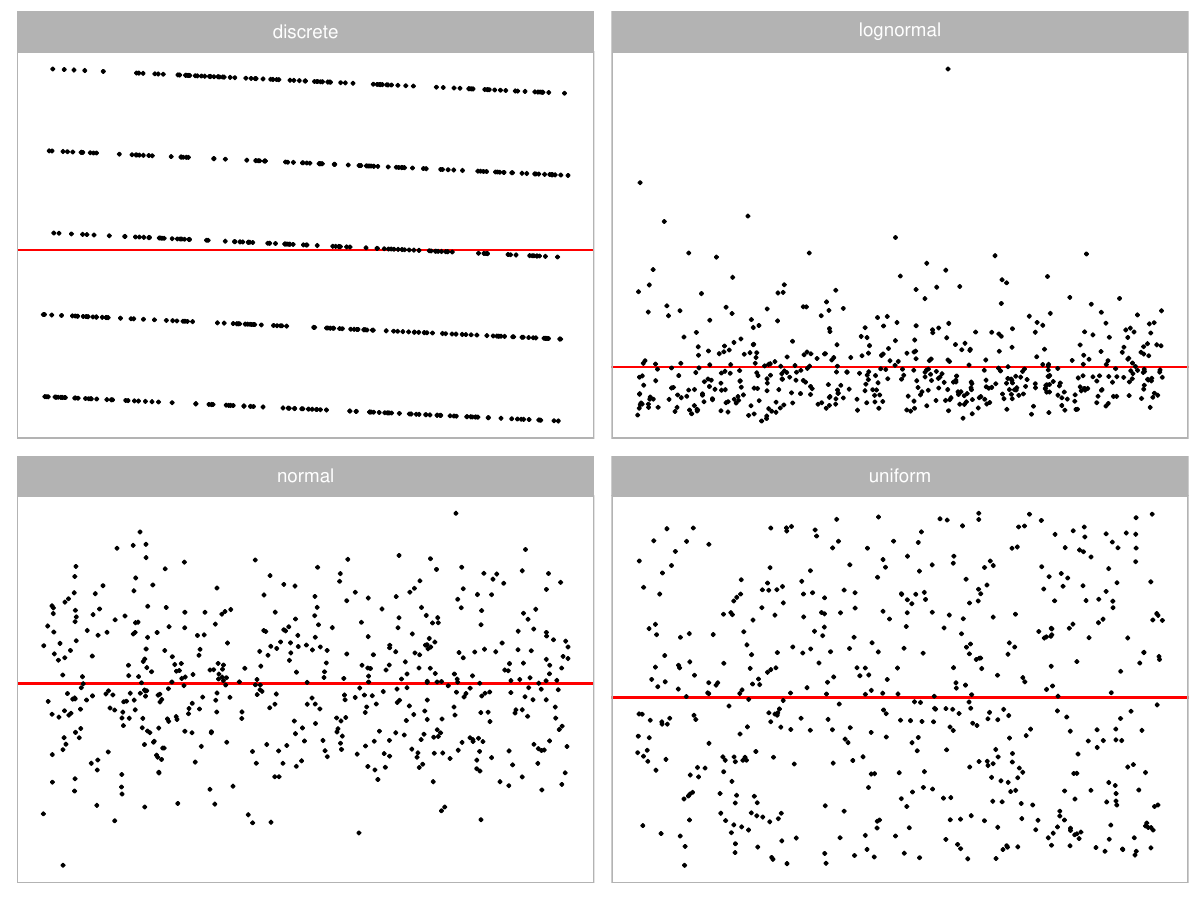} 

}

\caption{Non-normality forms generated for the synthetic data simulation. Four different error distributions including discrete, lognormal, normal and uniform are considered.}\label{fig:different-e}
\end{figure}

Non-normality violations arise from specifying a non-normal distribution
for \(\boldsymbol{\varepsilon}\). In the synthetic data simulation, four
distinct error distributions are considered, including discrete,
uniform, normal, and lognormal distributions, as presented in Figure
\ref{fig:different-e}. Each distribution imparts unique characteristics
in the residual plot. The discrete error distribution introduces
discrete clusters in residuals, while the lognormal distribution
typically yields outliers. Uniform error distribution may result in
residuals filling the entire space of the residual plot. All of these
distributions exhibit visual distinctions from the normal error
distribution.

\begin{figure}[!h]

{\centering \includegraphics[width=1\linewidth]{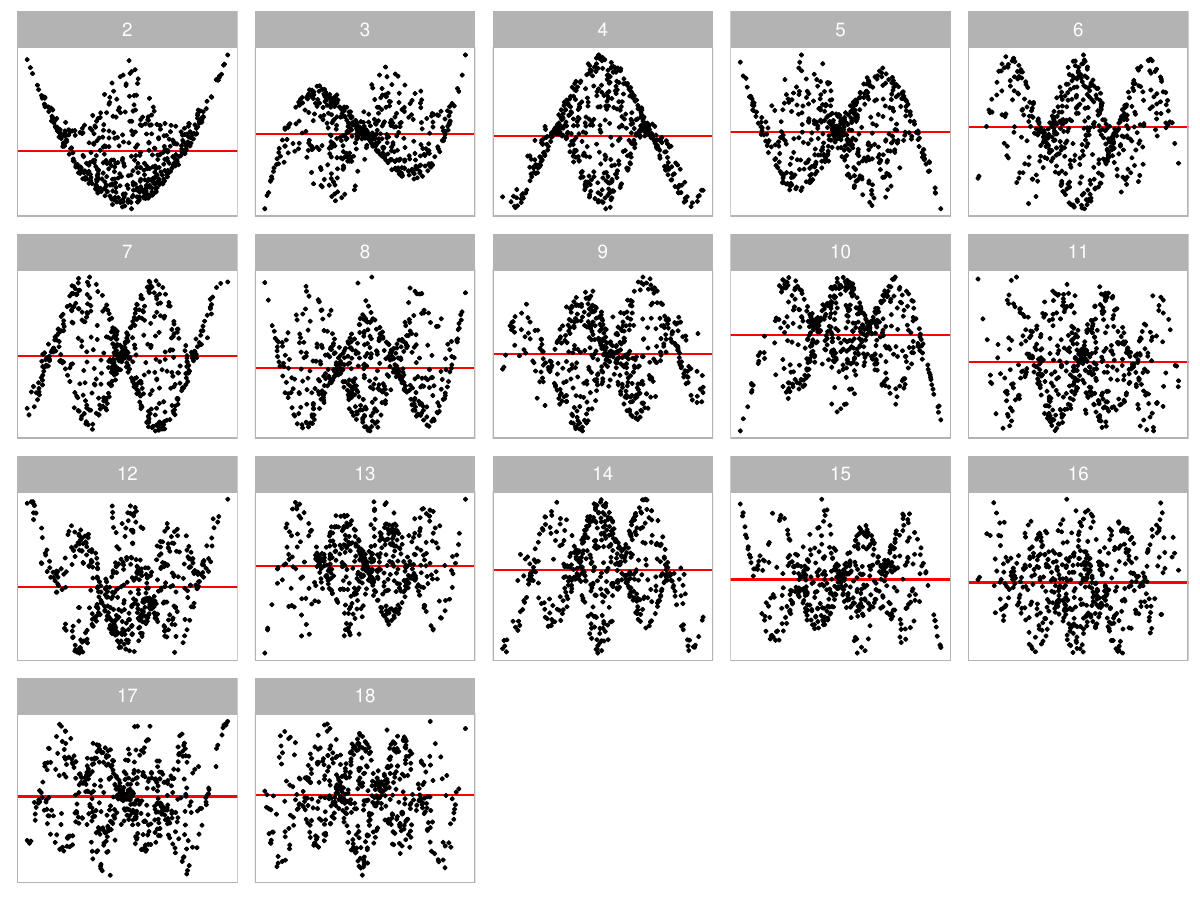} 

}

\caption{Residual plots of multiple linear regression models with non-linearity issues. The 17 shapes are generated by varying the order of polynomial given by $j$ in $He_j(.)$. A second predictor $\boldsymbol{x}_2$ is introduced to the regression model to create complex shapes.}\label{fig:different-j-x2}
\end{figure}

\begin{figure}[!h]

{\centering \includegraphics[width=1\linewidth]{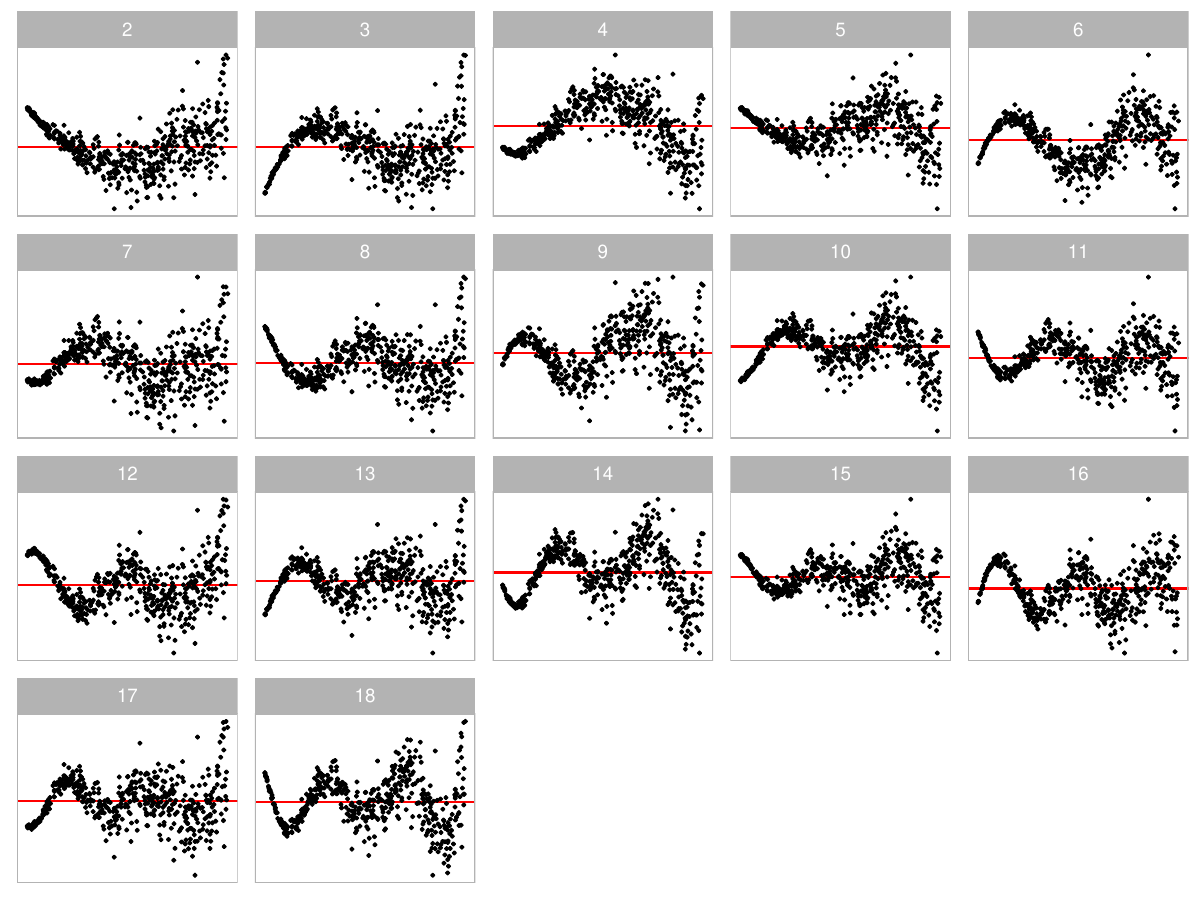} 

}

\caption{Residual plots of models violating both the non-linearity and the heteroskedasticity assumptions. The 17 shapes are generated by varying the order of polynomial given by $j$ in $He_j(.)$, and the "left-triangle" shape is introduced by setting $a = -1$.}\label{fig:different-j-heter}
\end{figure}

\begin{figure}[!h]

{\centering \includegraphics[width=1\linewidth]{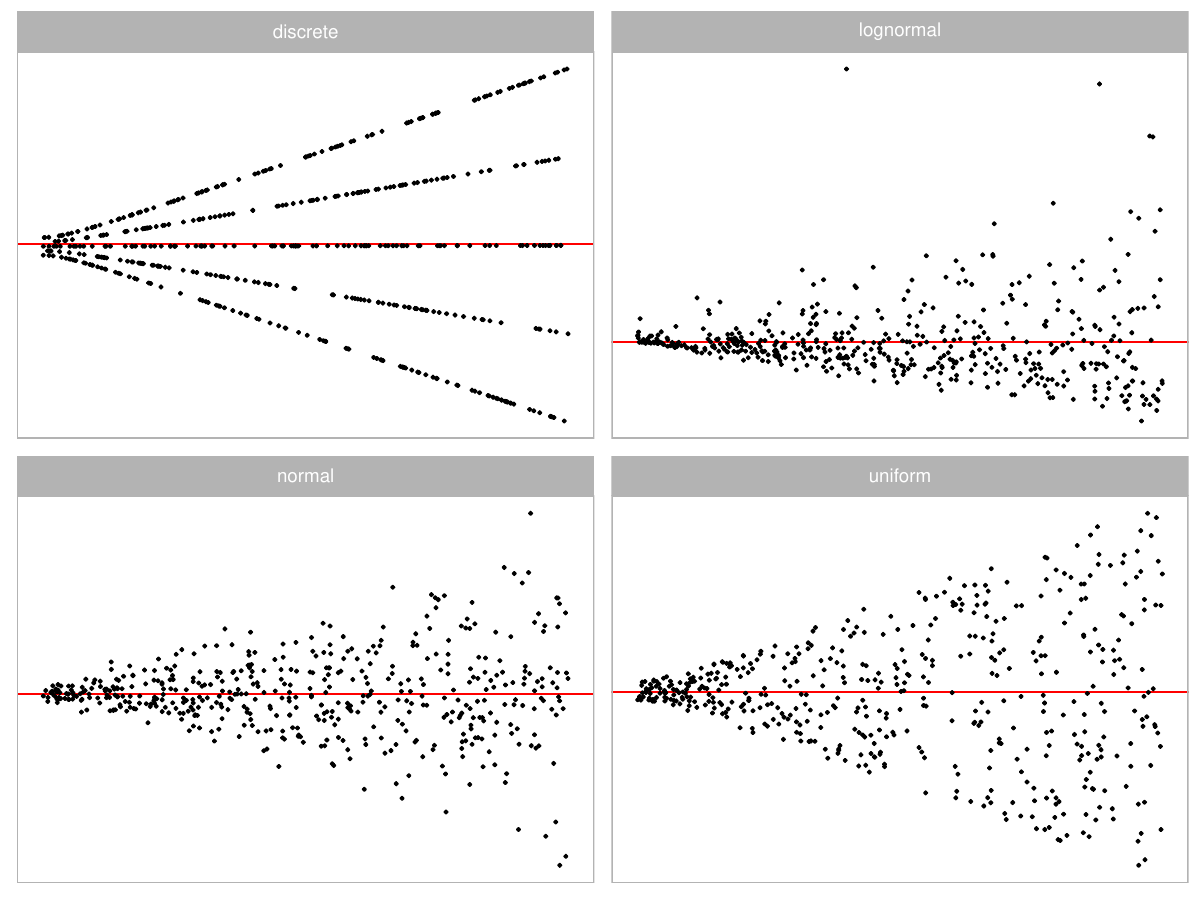} 

}

\caption{Residual plots of models violating both the non-normality and the heteroskedasticity assumptions. The four shapes are generated by using four different error distributions including discrete, lognormal, normal and uniform, and the "left-triangle" shape is introduced by setting $a = -1$. }\label{fig:different-e-heter}
\end{figure}

Equation \ref{eq:data-sim} accommodates the incorporation of the second
predictor \(\boldsymbol{x}_2\). Introducing it into the data generation
process by setting \(\beta_1 = 1\) significantly enhances the complexity
of the shapes, as illustrated in Figure \ref{fig:different-j-x2}. In
comparison to Figure \ref{fig:different-j}, Figure
\ref{fig:different-j-x2} demonstrates that the non-linear shape
resembles a surface rather than a single curve. This augmentation can
facilitate the computer vision model in learning visual patterns from
residual plots of the multiple linear regression model.

In real-world analysis, it is not uncommon to encounter instances where
multiple model violations coexist. In such cases, the residual plots
often exhibit a mixed pattern of visual anomalies corresponding to
different types of model violations. Figure \ref{fig:different-j-heter}
and Figure \ref{fig:different-e-heter} show the visual patterns of
models with multiple model violations.

\subsection{Balanced Dataset}\label{balanced-dataset}

To train a robust computer vision model, we deliberately controlled the
distribution of the target variable \(D\) in the training data. We
ensured that it followed a uniform distribution between \(0\) and \(7\).
This was achieved by organizing \(50\) buckets, each exclusively
accepting training samples with \(D\) falling within the range
\([7(i - 1)/49, 7i/49)\) for \(i < 50\), where \(i\) represents the
index of the \(i\)-th bucket. For the \(50\)-th bucket, any training
samples with \(D \geq 7\) were accepted.

With 80,000 training images prepared, each bucket accommodated a maximum
of \(80000/ 50 = 1600\) training samples. The simulator iteratively
sampled parameter values from the parameter space, generated residuals
and fitted values using the data generation process, computed the
distance, and checked if the sample fitted within the corresponding
bucket. This process continued until all buckets were filled.

Similarly, we adopted the same methodology to prepare 8,000 test images
for performance evaluation and model diagnostics.

\section{Model Architecture}\label{sec-model-architecture}

\begin{figure}

{\centering \includegraphics[width=1\linewidth]{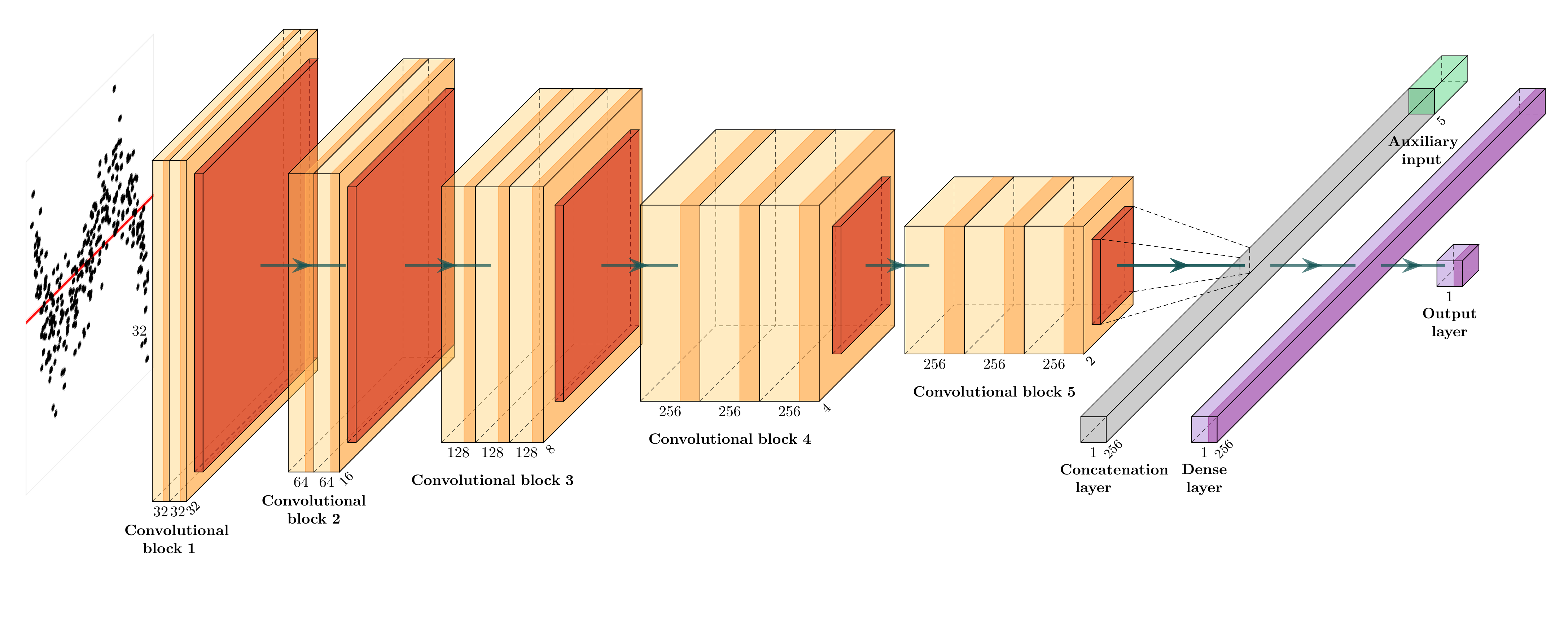} 

}

\caption{Diagram of the architecture of the optimized computer vision model. Numbers at the bottom of each box show the shape of the output of each layer. The band of each box drawn in a darker color indicates the use of the rectified linear unit activation function.  Yellow boxes are 2D convolutional layers, orange boxes are pooling layers, the grey box is the concatenation layer, and the purple boxes are dense layers.}\label{fig:cnn-diag}
\end{figure}

The architecture of the computer vision model is adapted from the
well-established VGG16 architecture \citep{simonyan2014very}. While more
recent architectures like ResNet \citep{he2016deep} and
DenseNet\citep{huang2017densely}, have achieved even greater
performance, VGG16 remains a solid choice for many applications due to
its simplicity and effectiveness. Our decision to use VGG16 aligns with
our goal of starting with a proven and straightforward model. Figure
\ref{fig:cnn-diag} provides a diagram of the architecture. More details
about the neural network layers used in this study are provided in
Appendix A.

The model begins with an input layer of shape
\(n \times h \times w \times 3\), capable of handling \(n\) RGB images.
This is followed by a grayscale conversion layer utilizing the luma
formula under the Rec. 601 standard \citep{series2011studio}, which
converts the color image to grayscale. Grayscale suffices for our task
since data points are plotted in black. We experiment with three
combinations of \(h\) and \(w\): \(32 \times 32\), \(64 \times 64\), and
\(128 \times 128\), aiming to achieve sufficiently high image resolution
for the problem at hand.

The processed image is used as the input for the first convolutional
block. The model comprises at most five consecutive convolutional
blocks, mirroring the original VGG16 architecture. Within each block,
there are two 2D convolutional layers followed by two activation layers,
respectively. Subsequently, a 2D max-pooling layer follows the second
activation layer. The 2D convolutional layer convolves the input with a
fixed number of \(3 \times 3\) convolution filters, while the 2D
max-pooling layer downsamples the input along its spatial dimensions by
taking the maximum value over a \(2 \times 2\) window for each channel
of the input. The activation layer employs the rectified linear unit
(ReLU) activation function, a standard practice in deep learning, which
introduces a non-linear transformation of the output of the 2D
convolutional layer. Additionally, to regularize training, a batch
normalization layer is added after each 2D convolutional layer and
before the activation layer. Finally, a dropout layer is appended at the
end of each convolutional block to randomly set some inputs to zero
during training, further aiding in regularization.

The output of the last convolutional block is summarized by either a
global max pooling layer or a global average pooling layer, resulting in
a two-dimensional tensor. To leverage the information contained in
scagnostics, this tensor is concatenated with an additional
\(n \times 5\) tensor, which contains the ``Monotonic'', ``Sparse'',
``Splines'', and ``Striped'' measures, along with the number of
observations for \(n\) residual plots.

The concatenated tensor is then fed into the final prediction block.
This block consists of two fully-connected layers. The first layer
contains at least \(128\) units, followed by a dropout layer.
Occasionally, a batch normalization layer is inserted between the
fully-connected layer and the dropout layer for regularization purposes.
The second fully-connected layer consists of only one unit, serving as
the output of the model.

The model weights \(\boldsymbol{\theta}\) were randomly initialized and
they were optimized by the Adam optimizer \citep{kingma2014adam} with
the mean square error loss function

\[\hat{\boldsymbol{\theta}} = \underset{\boldsymbol{\theta}}{\text{arg min}}\frac{1}{n_{\text{train}}}\sum_{i=1}^{n_{\text{train}}}(D_i - f_{\boldsymbol{\theta}}(V_i, S_i))^2,\]

\noindent where \(n_{\text{train}}\) is the number of training samples,
\(V_i\) is the \(i\)-th residual plot and \(S_i\) is the additional
information about the \(i\)-th residual plot including four scagnostics
and the number of observations.

\section{Model Training}\label{sec-model-training}

To achieve a near-optimal deep learning model, hyperparameters like the
learning rate often need to be fine-tuned using a tuner. In our study,
we utilized the Bayesian optimization tuner from the \texttt{KerasTuner}
Python library \citep{omalley2019kerastuner} for this purpose. A
comprehensive list of hyperparameters is provided in Table
\ref{tab:hyperparameter}.

The number of base filters determines the number of filters for the
first 2D convolutional layer. In the VGG16 architecture, the number of
filters for the 2D convolutional layer in a block is typically twice the
number in the previous block, except for the last block, which maintains
the same number of convolution filters as the previous one. This
hyperparameter aids in controlling the complexity of the computer vision
model. A higher number of base filters results in more trainable
parameters. Likewise, the number of units for the fully-connected layer
determines the complexity of the final prediction block. Increasing the
number of units enhances model complexity, resulting in more trainable
parameters.

The dropout rate and batch normalization are flexible hyperparameters
that work in conjunction with other parameters to facilitate smooth
training. A higher dropout rate is necessary when the model tends to
overfit the training dataset by learning too much noise
\citep{srivastava2014dropout}. Conversely, a lower dropout rate is
preferred when the model is complex and challenging to converge. Batch
normalization, on the other hand, addresses the internal covariate shift
problem arising from the randomness in weight initialization and input
data \citep{goodfellow2016deep}. It helps stabilize and accelerate the
training process by normalizing the activations of each layer.

Additionally, incorporating additional inputs such as scagnostics and
the number of observations can potentially enhance prediction accuracy.
Therefore, we allow the tuner to determine whether these inputs were
necessary for optimal model performance.

The learning rate is a crucial hyperparameter, as it dictates the step
size of the optimization algorithm. A high learning rate can help the
model avoid local minima but risks overshooting and missing the global
minimum. Conversely, a low learning rate smoothens the training process
but makes the convergence time longer and increases the likelihood of
getting trapped in local minima.

Our model was trained on the MASSIVE M3 high-performance computing
platform \citep{goscinski2014multi}, using TensorFlow
\citep{abadi2016tensorflow} and Keras \citep{chollet2015keras}. During
training, 80\% of the training data was utilized for actual training,
while the remaining 20\% was used as validation data. The Bayesian
optimization tuner conducted 100 trials to identify the best
hyperparameter values based on validation root mean square error. The
tuner then restored the best epoch of the best model from the trials.
Additionally, we applied early stopping, terminating the training
process if the validation root mean square error fails to improve for 50
epochs. The maximum allowed epochs was set at 2,000, although no models
reached this threshold.

\begin{table}

\caption{\label{tab:hyperparameter}Name of hyperparameters and their correspoding domain for the computer vision model.}
\centering
\begin{tabular}[t]{ll}
\toprule
Hyperparameter & Domain\\
\midrule
Number of base filters & \{4, 8, 16, 32, 64\}\\
Dropout rate for convolutional blocks & {}[0.1, 0.6]\\
Batch normalization for convolutional blocks & \{false, true\}\\
Type of global pooling & \{max, average\}\\
Ignore additional inputs & \{false, true\}\\
\addlinespace
Number of units for the fully-connected layer & \{128, 256, 512, 1024, 2048\}\\
Batch normalization for the fully-connected layer & \{false, true\}\\
Dropout rate for the fully-connected layer & {}[0.1, 0.6]\\
Learning rate & {}[$10^{-8}$, $10^{-1}$]\\
\bottomrule
\end{tabular}
\end{table}

Based on the tuning process described above, the optimized
hyperparameter values are presented in Table
\ref{tab:best-hyperparameter}. It was observable that a minimum of
\(32\) base filters was necessary, with the preferable choice being
\(64\) base filters for both the \(64 \times 64\) and \(128 \times 128\)
models, mirroring the original VGG16 architecture. The optimized dropout
rate for convolutional blocks hovered around \(0.4\), and incorporating
batch normalization for convolutional blocks proved beneficial for
performance.

All optimized models chose to retain the additional inputs, contributing
to the reduction of validation error. The number of units required for
the fully-connected layer was \(256\), a relatively modest number
compared to the VGG16 classifier, suggesting that the problem at hand
was less complex. The optimized learning rates were higher for models
with higher resolution input, likely because models with more parameters
are more prone to getting stuck in local minima, requiring a higher
learning rate.

\begin{table}

\caption{\label{tab:best-hyperparameter}Hyperparameters values for the optimized computer vision models with different input sizes.}
\centering
\resizebox{\linewidth}{!}{
\begin{tabular}[t]{llll}
\toprule
Hyperparameter & $32 \times 32$ & $64 \times 64$ & $128 \times 128$\\
\midrule
Number of base filters & 32 & 64 & 64\\
Dropout rate for convolutional blocks & 0.4 & 0.3 & 0.4\\
Batch normalization for convolutional blocks & true & true & true\\
Type of global pooling & max & average & average\\
Ignore additional inputs & false & false & false\\
\addlinespace
Number of units for the fully-connected layer & 256 & 256 & 256\\
Batch normalization for the fully-connected layer & false & true & true\\
Dropout rate for the fully-connected layer & 0.2 & 0.4 & 0.1\\
Learning rate & 0.0003 & 0.0006 & 0.0052\\
\bottomrule
\end{tabular}}
\end{table}

\section{Results}\label{sec-model-results}

\subsection{Model Performance}\label{model-performance}

The test performance for the optimized models with three different input
sizes are summarized in Table \ref{tab:performance}. Among these models,
the \(32 \times 32\) model consistently exhibited the best test
performance. The mean absolute error of the \(32 \times 32\) model
indicated that the difference between \(\hat{D}\) and \(D\) was
approximately \(0.43\) on the test set, a negligible deviation
considering the normal range of \(D\) typically falls between \(0\) and
\(7\). The high \(R^2\) values also suggested that the predictions were
largely linearly correlated with the target.

\begin{figure}[!h]

{\centering \includegraphics[width=1\linewidth]{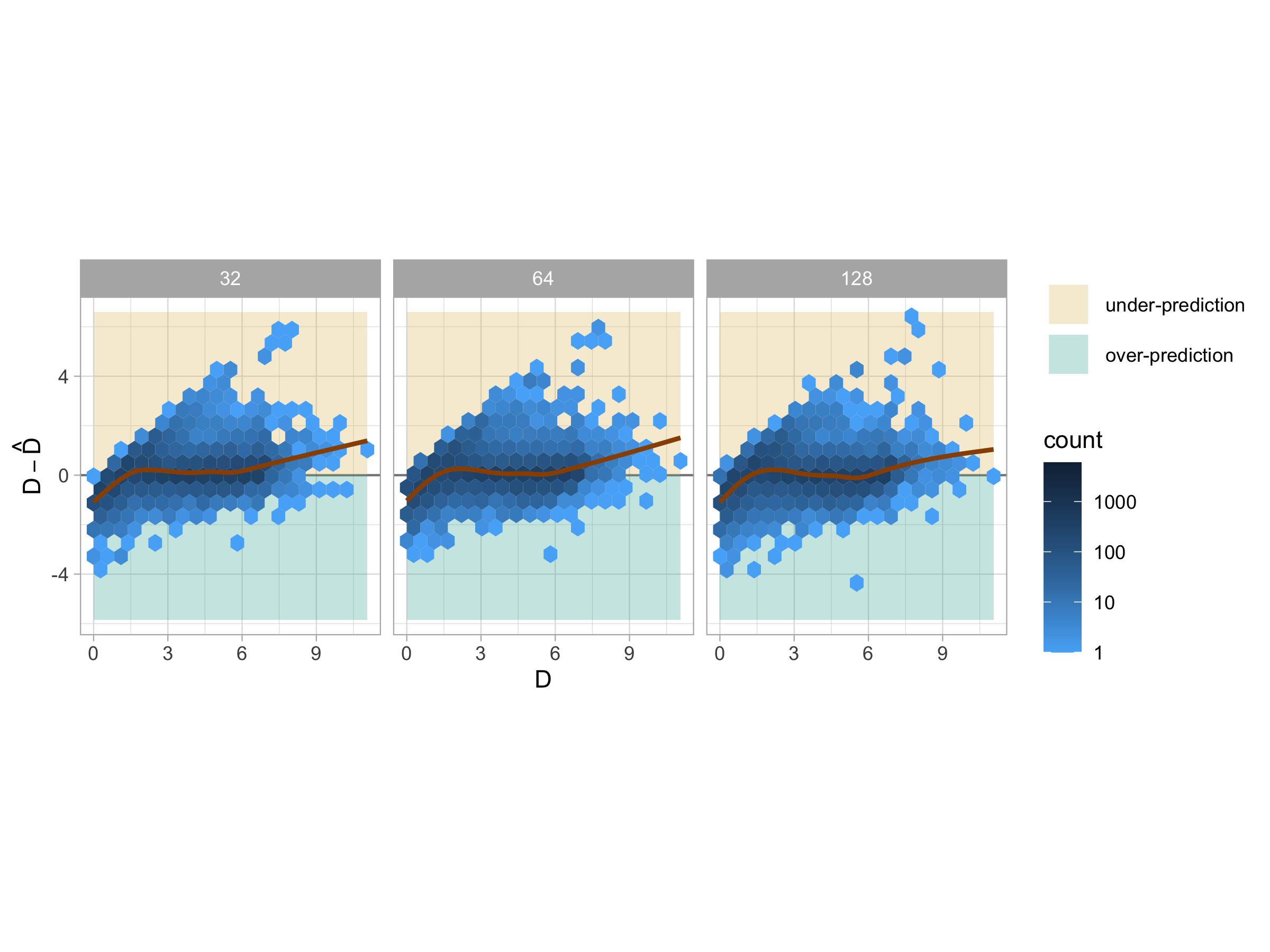} 

}

\caption{Hexagonal heatmap for difference in $D$ and $\hat{D}$ vs $D$ on test data for three optimized models with different input sizes. The brown lines are smoothing curves produced by fitting generalized additive models. The area over the zero line in light yellow indicates under-prediction, and the area under the zero line in light green indicates over-prediction.}\label{fig:model-performance}
\end{figure}

Figure \ref{fig:model-performance} presents a hexagonal heatmap for
\(D - \hat{D}\) versus \(D\). The brown smoothing curves, fitted by
generalized additive models \citep{hastie2017generalized}, demonstrate
that all the optimized models perform admirably on the test sets when
\(1.5 < D < 6\), where no structural issues are noticeable. However,
over-predictions occurred when \(D < 1.5\), while under-predictions
occurred predominantly when \(\hat{D} > 6\).

\begin{figure}[!h]

{\centering \includegraphics[width=1\linewidth]{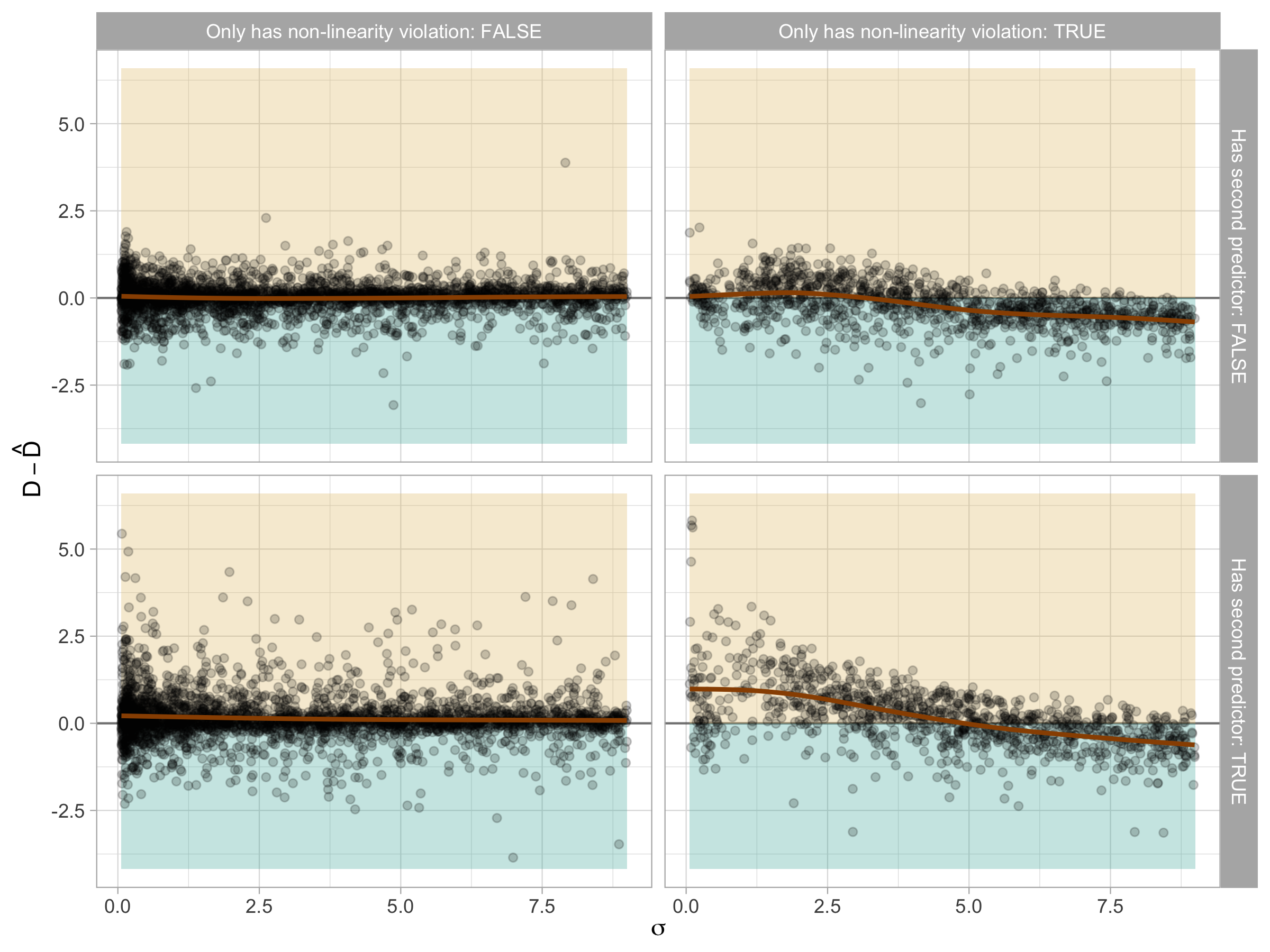} 

}

\caption{Scatter plots for difference in $D$ and $\hat{D}$ vs $\sigma$ on test data for the $32 \times 32$ optimized model. The data is grouped by whether the regression has only non-linearity violation, and whether it includes a second predictor in the regression formula. The brown lines are smoothing curves produced by fitting generalized additive models. The area over the zero line in light yellow indicates under-prediction, and the area under the zero line in light green indicates over-prediction.}\label{fig:over-under}
\end{figure}

For input images representing null plots where \(D = 0\), it was
expected that the models will over-predict the distance, as explained in
Section \ref{sec-model-lineup-evaluation}. However, it can not explain
the under-prediction issue. Therefore, we analysed the relationship
between residuals and all the factors involved in the data generating
process. We found that most issues actually arose from non-linearity
problems and the presence of a second predictor in the regression model
as illustrated in Figure \ref{fig:over-under}. When the variance for the
error distribution was small, the optimized model tended to
under-predict the distance. Conversely, when the error distribution had
a large variance, the model tended to over-predict the distance.

Since most of the deviation stemmed from the presence of non-linearity
violations, to further investigate this, we split the test set based on
violation types and re-evaluated the performance, as detailed in Table
\ref{tab:performance-sub}. It was evident that metrics for null plots
were notably worse compared to other categories. Furthermore, residual
plots solely exhibiting non-normality issues were the easiest to
predict, with very low test root mean square error (RMSE) at around
\(0.3\). Residual plots with non-linearity issues were more challenging
to assess than those with heteroskedasticity or non-normality issues.
When multiple violations were introduced to a residual plot, the
performance metrics typically lay between the metrics for each
individual violation.

Based on the model performance metrics, we chose to use the
best-performing model evaluated on the test set, namely the
\(32 \times 32\) model, for the subsequent analysis.

\begin{table}

\caption{\label{tab:performance}The test performance of three optimized models with different input sizes.}
\centering
\begin{tabular}[t]{lrrrr}
\toprule
 & RMSE & $R^2$ & MAE & Huber loss\\
\midrule
$32 \times 32$ & 0.660 & 0.901 & 0.434 & 0.18\\
$64 \times 64$ & 0.674 & 0.897 & 0.438 & 0.19\\
$128 \times 128$ & 0.692 & 0.892 & 0.460 & 0.20\\
\bottomrule
\end{tabular}
\end{table}

\begin{table}

\caption{\label{tab:performance-sub}The training and test performance of the $32 \times 32$ model presented with different model violations.}
\centering
\begin{tabular}[t]{lrr}
\toprule
Violations & \#samples & RMSE\\
\midrule
no violations & 155 & 1.267\\
non-linearity & 2218 & 0.787\\
heteroskedasticity & 1067 & 0.602\\
non-linearity + heteroskedasticity & 985 & 0.751\\
non-normality & 1111 & 0.320\\
non-linearity + non-normality & 928 & 0.600\\
heteroskedasticity + non-normality & 819 & 0.489\\
non-linearity + heteroskedasticity + non-normality & 717 & 0.620\\
\bottomrule
\end{tabular}
\end{table}

\subsection{Comparison with Human Visual Inference and Conventional
Tests}\label{comparison-with-human-visual-inference-and-conventional-tests}

\subsubsection{Overview of the Human Subject
Experiment}\label{overview-of-the-human-subject-experiment}

In order to check the validity of the proposed computer vision model,
residual plots presented in the human subject experiment conducted by
\citet{li2024plot} will be assessed.

This study has collected 7,974 human responses to 1,152 lineups. Each
lineup contains one randomly placed true residual plot and 19 null
plots. Among the 1,152 lineups, 24 are attention check lineups in which
the visual patterns are designed to be extremely obvious and very
different from the corresponding to null plots, 36 are null lineups
where all the lineups consist of only null plots, 279 are lineups with
uniform predictor distribution evaluated by 11 participants, and the
remaining 813 are lineups with discrete, skewed or normal predictor
distribution evaluated by 5 participants. Attention check lineups and
null lineups will not be assessed in the following analysis.

In \citet{li2024plot}, the residual plots are simulated from a data
generating process which is a special case of Equation
\ref{eq:data-sim}. The main characteristic is the model violations are
introduced separately, meaning non-linearity and heteroskedasticity will
not co-exist in one lineup but assigned uniformly to all lineups.
Additionally, non-normality and multiple predictors are not considered
in the experimental design.

\subsubsection{Model Performance on the Human-evaluated
Data}\label{model-performance-on-the-human-evaluated-data}

\begin{table}

\caption{\label{tab:experiment-performance}The performance of the $32 \times 32$ model on the data used in the human subject experiment.}
\centering
\begin{tabular}[t]{lrrrr}
\toprule
Violation & RMSE & $R^2$ & MAE & Huber loss\\
\midrule
heteroskedasticity & 0.721 & 0.852 & 0.553 & 0.235\\
non-linearity & 0.738 & 0.770 & 0.566 & 0.246\\
\bottomrule
\end{tabular}
\end{table}

For each lineup used in \citet{li2024plot}, there is one true residual
plot and 19 null plots. While the distance \(D\) for the true residual
plot depends on the underlying data generating process, the distance
\(D\) for the null plots is zero. We have used our optimized computer
vision model to estimate distance for both the true residual plots and
the null plots. To have a fair comparison, \(H_0\) will be rejected if
the true residual plot has the greatest estimated distance among all
plots in a lineup. Additionally, the appropriate conventional tests
including the Ramsey Regression Equation Specification Error Test
(RESET) \citep{ramsey1969tests} for non-linearity and the Breusch-Pagan
test \citep{breusch1979simple} for heteroskedasticity were applied on
the same data for comparison.

The performance metrics of \(\hat{D}\) for true residual plots are
outlined in Table \ref{tab:experiment-performance}. It is notable that
all performance metrics are slightly worse than those evaluated on the
test data. Nevertheless, the mean absolute error remains at a low level,
and the linear correlation between the prediction and the true value
remains very high. Consistent with results in Table
\ref{tab:performance-sub}, lineups with non-linearity issues are more
challenging to predict than those with heteroskedasticity issues.

\begin{figure}[!h]

{\centering \includegraphics[width=1\linewidth]{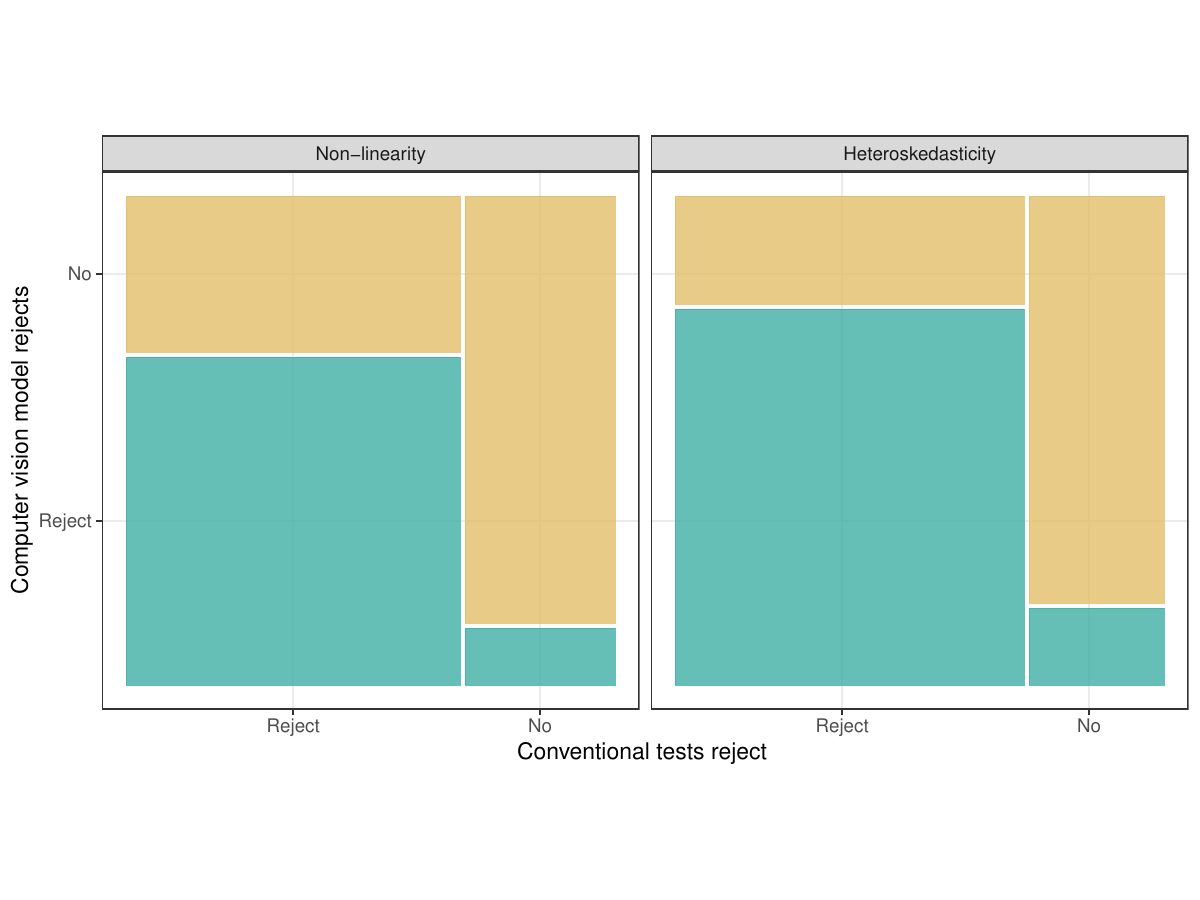} 

}

\caption{Rejection rate ($p$-value $\leq0.05$) of computer vision models conditional on conventional tests on non-linearity (left) and heteroskedasticity (right) lineups displayed using a mosaic plot. When the conventional test fails to reject, the computer vision mostly fails to reject the same plot as well as indicated by the height of the top right yellow rectangle, but there are non negliable amount of plots where the conventional test rejects but the computer vision model fails to reject as indicated by the width of the top left yellow rectangle.}\label{fig:conv-mosaic}
\end{figure}

Table \ref{tab:human-conv-table} provides a summary of the agreement
between decisions made by the computer vision model and conventional
tests. The agreement rates between conventional tests and the computer
vision model are 85.95\% and 79.69\% for residual plots containing
heteroskedasticity and non-linearity patterns, respectively. These
figures are higher than those calculated for visual tests conducted by
human, indicating that the computer vision model exhibits behavior more
akin to the best available conventional tests. However, Figure
\ref{fig:conv-mosaic} shows that the computer vision model does not
always reject when the conventional tests reject. And a small number of
plots will be rejected by computer vision model but not by conventional
tests. This suggests that conventional tests are more sensitive than the
computer vision model.

\begin{figure}

{\centering \includegraphics[width=1\linewidth]{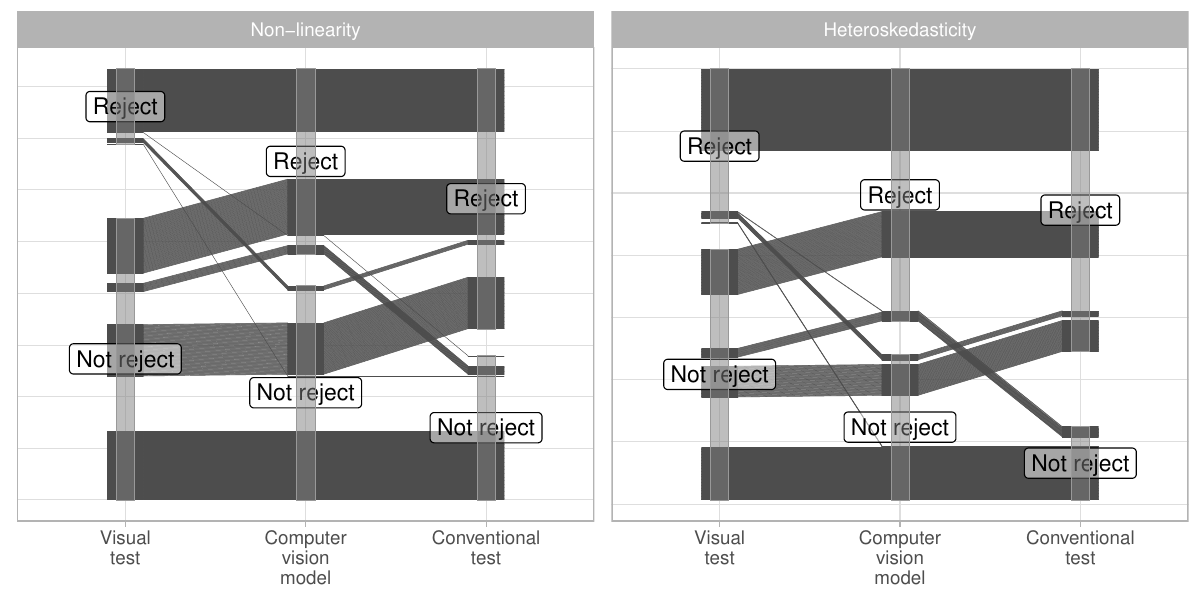} 

}

\caption{Parallel coordinate plots of decisions made by computer vision model, conventional tests and visual tests made by human.}\label{fig:pcp}
\end{figure}

Figure \ref{fig:pcp} further illustrates the decisions made by visual
tests conducted by human, computer vision models, and conventional
tests, using a parallel coordinate plots. It can be observed that all
three tests will agree with each other for around 50\% of the cases.
When visual tests conducted by human do not reject, there are
substantial amount of cases where computer vision model also do not
reject but conventional tests reject. There are much fewer cases that do
not reject by visual tests and conventional tests, but is rejected by
computer vision models. This indicates computer vision model can behave
like visual tests conducted by human better than conventional tests.
Moreover, there are great proportion of cases where visual tests
conducted by human is the only test who does not reject.

\begin{figure}[!h]

{\centering \includegraphics[width=1\linewidth]{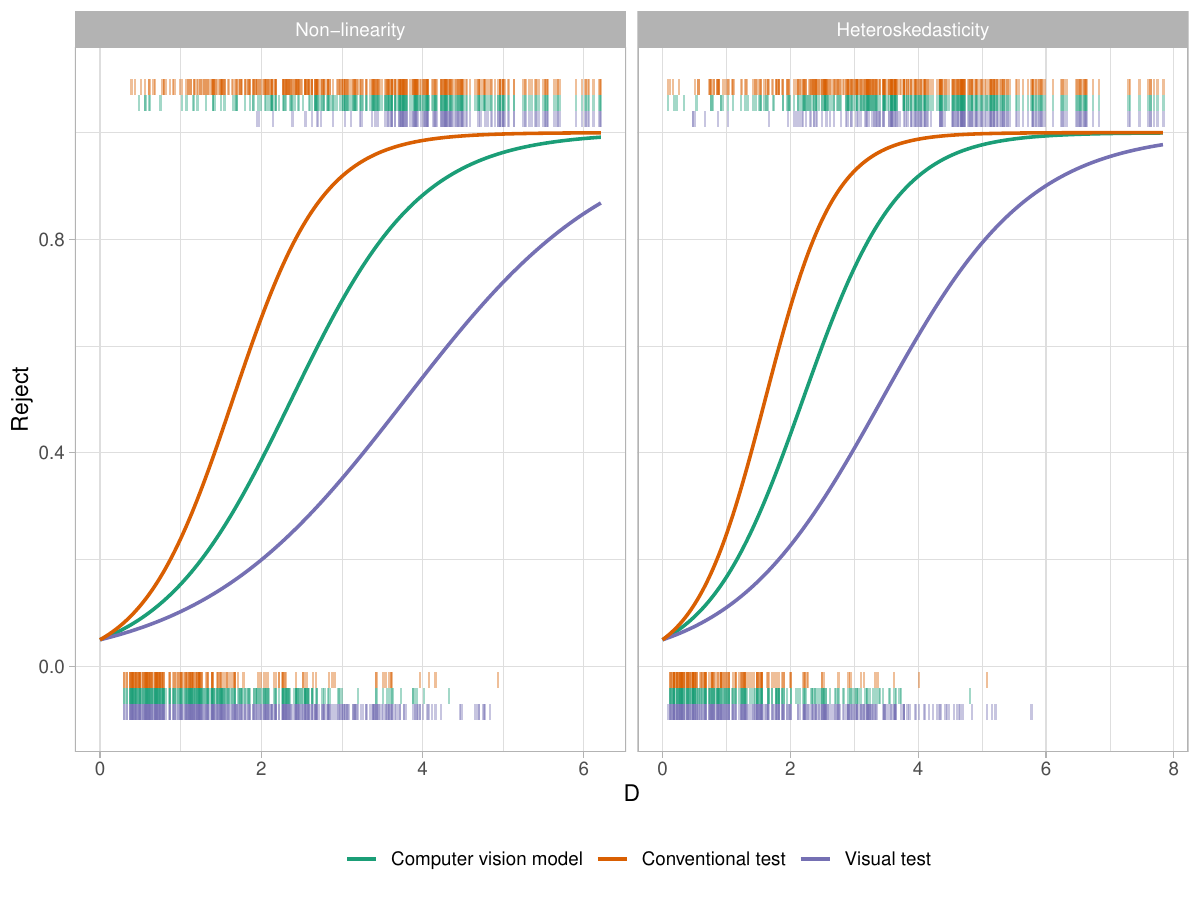} 

}

\caption{Comparison of power of visual tests, conventional tests and the computer vision model. Marks along the x-axis at the bottom of the plot represent rejections made by each type of test. Marks at the top of the plot represent acceptances. Power curves are fitted by logistic regression models with no intercept but an offset equals to $\text{log}(0.05/0.95)$.}\label{fig:power}
\end{figure}

When plotting the decision against the distance, as illustrated in
Figure \ref{fig:power}, several notable observations emerge. Firstly,
compared to conventional tests, the computer vision model tends to have
fewer rejected cases when \(D < 2\) and fewer non-rejected cases when
\(2< D < 4\). This suggests tests based on the computer vision model are
less sensitive to small deviations from model assumptions than
conventional tests but more sensitive to moderate deviations.
Additionally, visual tests demonstrate the lowest sensitivity to
residual plots with small distances where not many residual plots are
rejected when \(D < 2\). Similarly, for large distances where \(D > 4\),
almost all residual plots are rejected by the computer vision model and
conventional tests, but for visual tests conducted by humans, the
threshold is higher with \(D > 5\).

In Figure \ref{fig:power}, rejection decisions are fitted by logistic
regression models with no intercept terms and an offset equals to
\(\text{log}(0.05/0.95)\). The fitted curves for the computer vision
model fall between those of conventional tests and visual tests for both
non-linearity and heteroskedasticity, which means there is still
potential to refine the computer vision model to better align its
behavior with visual tests conducted by humans.

In the experiment conducted in \citet{li2024plot}, participants were
allowed to make multiple selections for a lineup. The weighted detection
rate was computed by assigning weights to each detection. If the
participant selected zero plots, a weight of 0.05 was assigned;
otherwise, if the true residual plot was detected, the weight was 1
divided by the number of selections. This weighted detection rate allow
us to assess the quality of the distance measure purposed in this paper,
by using the \(\delta\)-difference statistic. The \(\delta\)-difference
is originally defined by \citet{chowdhury2018measuring}, is given by

\[
\delta = \bar{d}_{\text{true}} - \underset{j}{\text{max}}\left(\bar{d}_{\text{null}}^{(j)}\right) \quad \text{for}~j = 1,...,m-1,
\]

where \(\bar{d}_{\text{null}}^{(j)}\) is the mean distance between the
\(j\)-th null plot and the other null plots, \(\bar{d}_{\text{true}}\)
is the mean distance between the true residual plot and null plots, and
\(m\) is the number of plots in a lineup. These mean distances are used
because, as noted by \citet{chowdhury2018measuring}, the distances can
vary depending on which data plot is used for comparison. For instance,
with three null plots, A, B and C, the distance between A and B may
differ from the distance between A and C. To obtain a consistent
distance for null plot A, averaging is necessary. However, this approach
is not applicable to the distance proposed in this paper, as we only
compare the residual plot against a theoretically good residual plot.
Consequently, the statistic must be adjusted to evaluate our distance
measure effectively.

One important aspect that the \(\delta\)-difference was designed to
capture is the empirical distribution of distances for null plot. If we
were to replace the mean distances \(\bar{d}_{\text{null}}^{(j)}\)
directly with \(D_{\text{null}}^{(j)}\), the distance of the \(j\)-th
null plot, the resulting distribution would be degenerate, since
\(D_{null}\) equals zero by definition. Additionally, \(D\) can not be
derived from an image, meaning it falls outside the scope of the
distances considered by \citet{chowdhury2018measuring}. Instead, the
focus should be on the empirical distribution of \(\hat{D}\), as it
influences decision-making. Therefore, the adjusted \(\delta\)-different
is defined as

\[
\delta_{\text{adj}} = \hat{D} - \underset{j}{\text{max}}\left(\hat{D}_{\text{null}}^{(j)}\right) \quad \text{for}~j = 1,...,m-1,
\]

\noindent where \(\hat{D}_{\text{null}}^{(j)}\) is the estimated
distance for the \(j\)-th null plot, and \(m\) is the number of plots in
a lineup.

\begin{figure}[!h]

{\centering \includegraphics[width=1\linewidth]{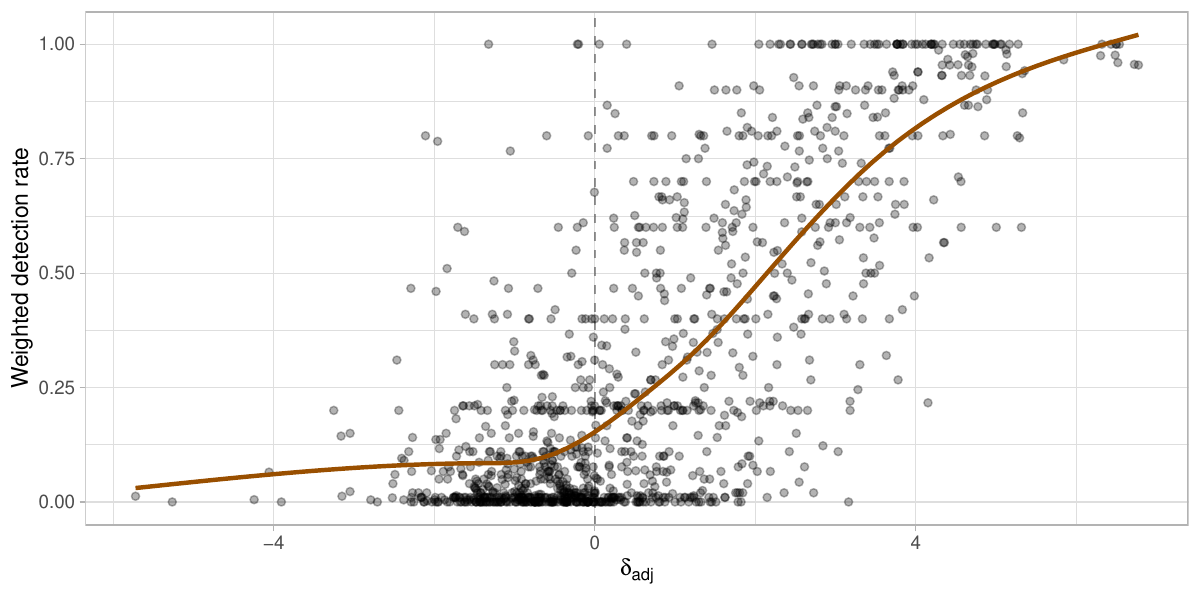} 

}

\caption{A weighted detection rate vs adjusted $\delta$-difference plot. The brown line is smoothing curve produced by fitting generalized additive models.}\label{fig:delta}
\end{figure}

Figure \ref{fig:delta} displays the scatter plot of the weighted
detection rate vs the adjusted \(\delta\)-difference. It indicates that
the weighted detection rate increases as the adjusted
\(\delta\)-difference increases, particularly when the adjusted
\(\delta\)-difference is greater than zero. A negative adjusted
\(\delta\)-difference suggests that there is at least one null plot in
the lineup with a stronger visual signal than the true residual plot. In
some instances, the weighted detection rate is close to one, yet the
adjusted \(\delta\)-difference is negative. This discrepancy implies
that the distance measure, or the estimated distance, may not perfectly
reflect actual human behavior.

\begin{table}

\caption{\label{tab:human-conv-table}Summary of the comparison of decisions made by computer vision model with decisions made by conventional tests and visual tests conducted by human.}
\centering
\begin{tabular}[t]{lrrr}
\toprule
Violations & \#Samples & \#Agreements & Agreement rate\\
\midrule
\addlinespace[0.3em]
\multicolumn{4}{l}{\textbf{Compared with conventional tests}}\\
\hspace{1em}heteroskedasticity & 540 & 464 & 0.8593\\
\hspace{1em}non-linearity & 576 & 459 & 0.7969\\
\addlinespace[0.3em]
\multicolumn{4}{l}{\textbf{Compared with visual tests conducted by human}}\\
\hspace{1em}heteroskedasticity & 540 & 367 & 0.6796\\
\hspace{1em}non-linearity & 576 & 385 & 0.6684\\
\bottomrule
\end{tabular}
\end{table}

\section{Examples}\label{sec-examples}

In this section, we present the performance of trained computer vision
model on three example datasets. These include the dataset associated
with the residual plot displaying a ``left-triangle'' shape, as
displayed in Figure \ref{fig:false-finding}, along with the Boston
housing dataset \citep{harrison1978hedonic}, and the ``dino'' datasets
from the \texttt{datasauRus} R package \citep{datasaurus}.

The first example illustrates a scenario where both the computer vision
model and human visual inspection successfully avoid rejecting \(H_0\)
when \(H_0\) is true, contrary to conventional tests. This underscores
the necessity of visually examining the residual plot.

In the second example, we encounter a more pronounced violation of the
model, resulting in rejection of \(H_0\) by all three tests. This
highlights the practicality of the computer vision model, particularly
for less intricate tasks.

The third example presents a situation where the model deviation is
non-typical. Here, the computer vision model and human visual inspection
reject \(H_0\), whereas some commonly used conventional tests do not.
This emphasizes the benefits of visual inspection and the unique
advantage of the computer vision model, which, like humans, makes
decisions based on visual discoveries.

\subsection{Left-triangle}\label{left-triangle}

\begin{figure}[!h]

{\centering \includegraphics[width=1\linewidth]{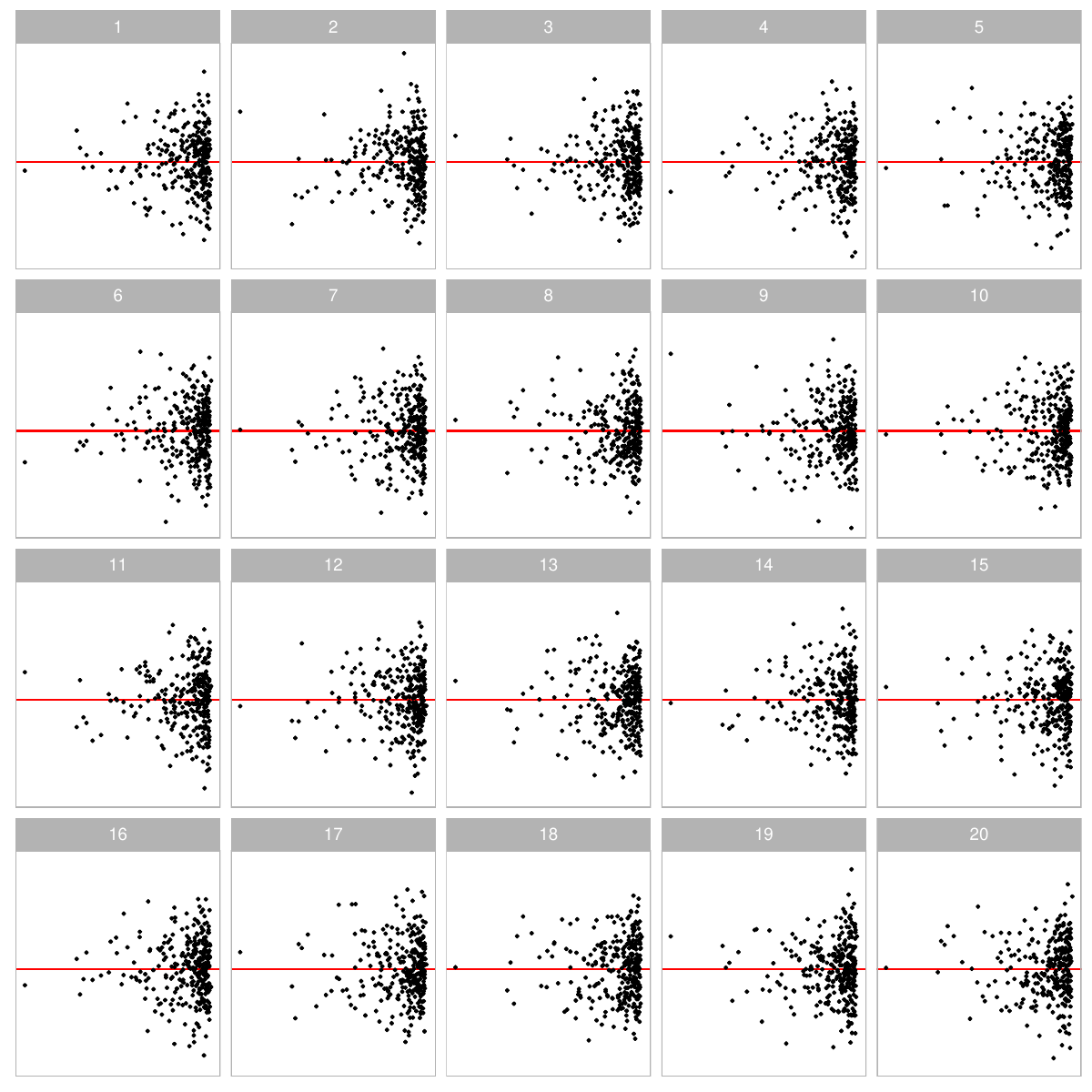} 

}

\caption{A lineup of residual plots displaying "left-triangle" visual patterns. The true residual plot occupies position 10, yet there are no discernible visual patterns that distinguish it from the other plots.}\label{fig:false-lineup}
\end{figure}

In Section \ref{sec-model-introduction}, we presented an example
residual plot showcased in Figure \ref{fig:false-finding}, illustrating
how humans might misinterpret the ``left-triangle'' shape as indicative
of heteroskedasticity. Additionally, the Breusch-Pagan test yielded a
rejection with a \(p\)-value of 0.046, despite the residuals originating
from a correctly specified model. Figure \ref{fig:false-lineup} offers a
lineup for this fitted model, showcasing various degrees of
``left-triangle'' shape across all residual plots. This phenomenon is
evidently caused by the skewed distribution of the fitted values.
Notably, if the residual plot were evaluated through a visual test, it
would not be rejected since the true residual plot positioned at 10 can
not be distinguished from the others.

\begin{figure}[!h]

{\centering \includegraphics[width=1\linewidth]{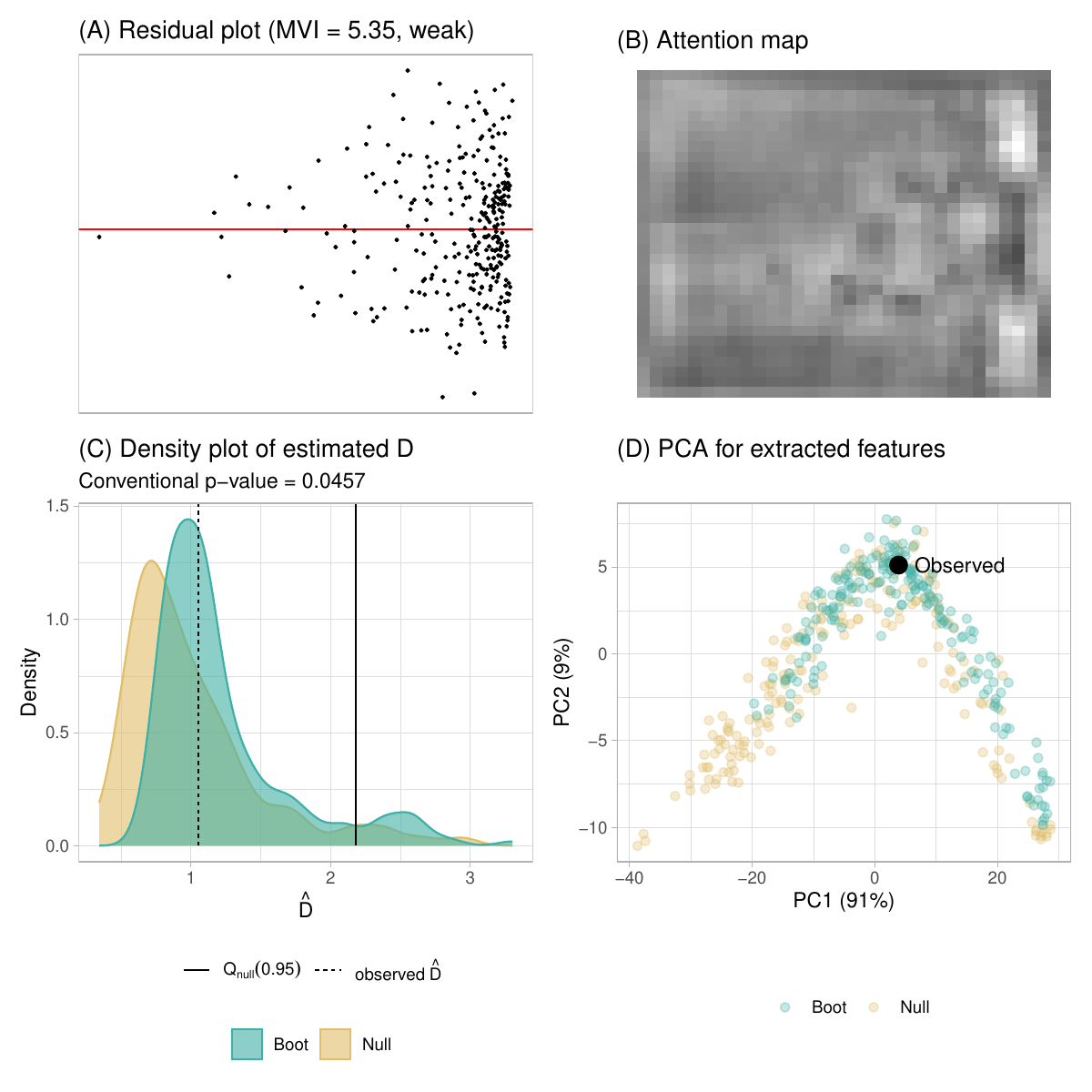} 

}

\caption{A summary of the residual plot assessment evaluted on 200 null plots and 200 bootstrapped plots. (A) The true residual plot exhibiting a "left-triangle" shape. (B) The attention map produced by computing the gradient of the output with respect to the greyscale input.  (C) The density plot of estimated distance for null plots and bootstrapped plots. The green area indicates the distribution of estimated distances for bootstrapped plots, while the yellow area represents the distribution of estimated distances for null plots. The fitted model will not be rejected since $\hat{D} < Q_{null}(0.95)$. (D) plot of first two principal components of features extracted from the global pooling layer of the computer vision model.  }\label{fig:false-check}
\end{figure}

Figure \ref{fig:false-check} presents the results of the assessment by
the computer vision model. Notably, the observed visual signal strength
is considerably lower than the 95\% sample quantile of the null
distribution. Moreover, the bootstrapped distribution suggests that it
is highly improbable for the fitted model to be misspecified as the
majority of bootstrapped fitted models will not be rejected. Thus, for
this particular fitted model, both the visual test and the computer
vision model will not reject \(H_0\). However, the Breusch-Pagan test
will reject \(H_0\) because it can not effectively utilize information
from null plots.

The attention map at Figure \ref{fig:false-check}B suggests that the
estimation is highly influenced by the top-right and bottom-right part
of the residual plot, as it forms two vertices of the triangular shape.
A principal component analysis (PCA) is also performed on the output of
the global pooling layer of the computer vision model. As mentioned in
\citet{simonyan2014very}, a computer vision model built upon the
convolutional blocks can be viewed as a feature extractor. For the
\(32 \times 32\) model, there are 256 features outputted from the global
pooling layer, which can be further used for different visual tasks not
limited to distance prediction. To see if these features can be
effectively used for distinguishing null plots and true residual plot,
we linearly project them into the first and second principal components
space as shown in Figure \ref{fig:false-check}D. It can be observed that
because the bootstrapped plots are mostly similar to the null plots, the
points drawn in different colors are mixed together. The true residual
plot is also covered by both the cluster of null plots and cluster of
bootstrapped plots. This accurately reflects our understanding of Figure
\ref{fig:false-lineup}.

\subsection{Boston Housing}\label{boston-housing}

The Boston housing dataset, originally published by
\citet{harrison1978hedonic}, offers insights into housing in the Boston,
Massachusetts area. For illustration purposes, we utilize a reduced
version from Kaggle, comprising 489 rows and 4 columns: average number
of rooms per dwelling (RM), percentage of lower status of the population
(LSTAT), pupil-teacher ratio by town (PTRATIO), and median value of
owner-occupied homes in \$1000's (MEDV). In our analysis, MEDV will
serve as the response variable, while the other columns will function as
predictors in a linear regression model. Our primary focus is to detect
non-linearity, because the relationships between RM and MEDV or LSTAT
and MEDV are non-linear.

\begin{figure}[!h]

{\centering \includegraphics[width=1\linewidth]{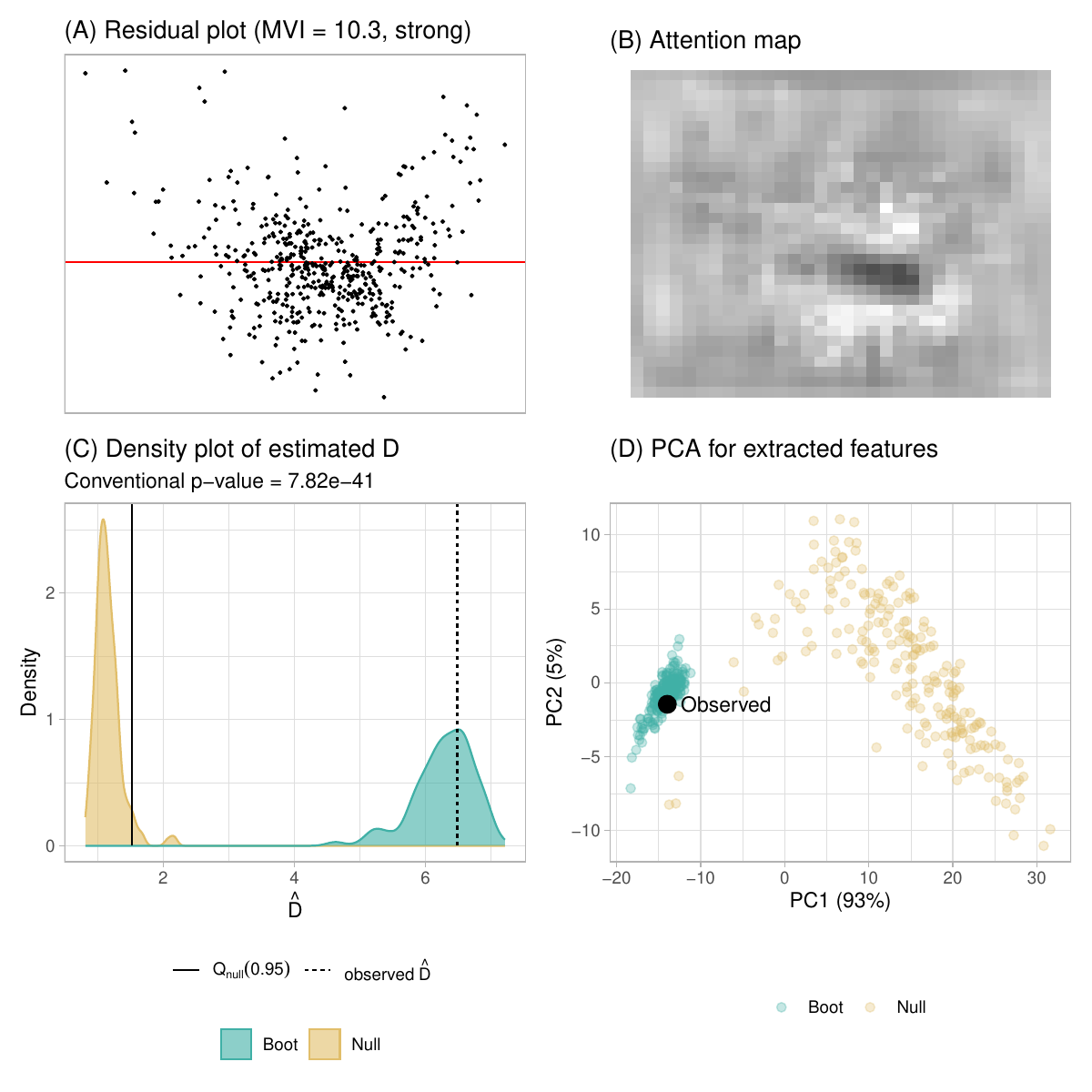} 

}

\caption{A summary of the residual plot assessment for the Boston housing fitted model evaluted on 200 null plots and 200 bootstrapped plots. (A) The true residual plot exhibiting a "U" shape. (B) The attention map produced by computing the gradient of the output with respect to the greyscale input.  (C) The density plot of estimated distance for null plots and bootstrapped plots. The blue area indicates the distribution of estimated distances for bootstrapped plots, while the yellow area represents the distribution of estimated distances for null plots. The fitted model will be rejected since $\hat{D} \geq Q_{null}(0.95)$. (D) plot of first two principal components of features extracted from the global pooling layer of the computer vision model. }\label{fig:boston-check}
\end{figure}

\begin{figure}[!h]

{\centering \includegraphics[width=1\linewidth]{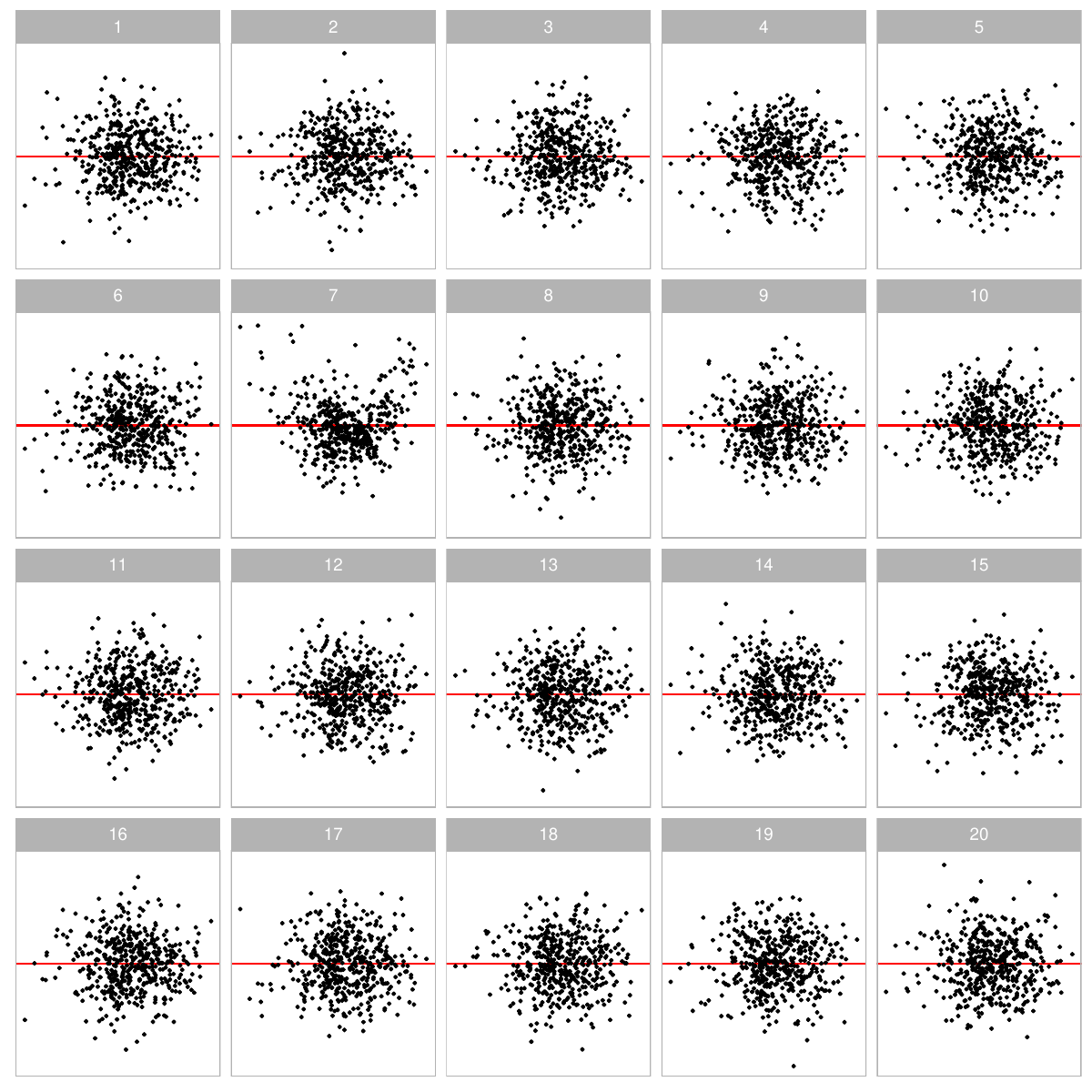} 

}

\caption{A lineup of residual plots for the Boston housing fitted model. The true residual plot is at position 7. It can be easily identified as the most different plot.}\label{fig:boston-lineup}
\end{figure}

Figure \ref{fig:boston-check} displays the residual plot and the
assessment conducted by the computer vision model. A clear non-linearity
pattern resembling a ``U'' shape is shown in the plot A. Furthermore,
the RESET test yields a very small \(p\)-value. The estimated distance
\(\hat{D}\) significantly exceeds \(Q_{null}(0.95)\), leading to
rejection of \(H_0\). The bootstrapped distribution also suggests that
almost all the bootstrapped fitted models will be rejected, indicating
that the fitted model is unlikely to be correctly specified. The
attention map in plot B suggests the center of the image has higher
leverage than other areas, and it is the turning point of the ``U''
shape. The CPA provided in plot D shows two distinct clusters of data
points, further underling the visual differences between bootstrapped
plots and null plots. This coincides the findings from Figure
\ref{fig:boston-lineup}, where the true plot exhibiting a ``U'' shape is
visually distinctive from null plots. If a visual test is conducted by
human, \(H_0\) will also be rejected.

\subsection{DatasauRus}\label{datasaurus}

The computer vision model possesses the capability to detect not only
typical issues like non-linearity, heteroskedasticity, and non-normality
but also artifact visual patterns resembling real-world objects, as long
as they do not appear in null plots. These visual patterns can be
challenging to categorize in terms of model violations. Therefore, we
will employ the RESET test, the Breusch-Pagan test, and the Shapiro-Wilk
test \citep{shapiro1965analysis} for comparison.

\begin{figure}[!h]

{\centering \includegraphics[width=1\linewidth]{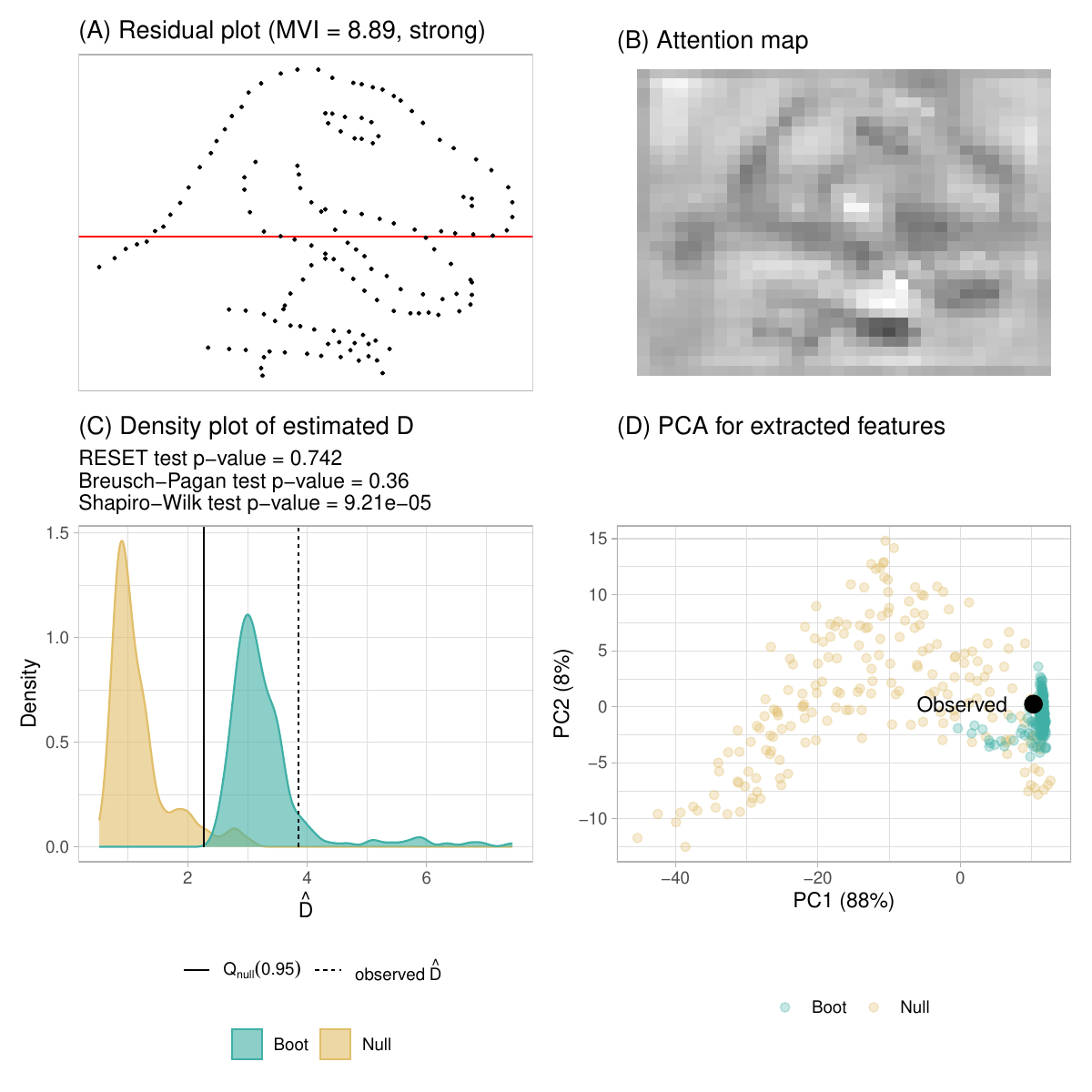} 

}

\caption{A summary of the residual plot assessment for the datasauRus fitted model evaluated on 200 null plots and 200 bootstrapped plots. (A) The residual plot exhibits a "dinosaur" shape. (B) The attention map produced by computing the gradient of the output with respect to the greyscale input.  (C) The density plot of estimated distance for null plots and bootstrapped plots. The blue area indicates the distribution of estimated distances for bootstrapped plots, while the yellow area represents the distribution of estimated distances for null plots. The fitted model will be rejected since $\hat{D} \geq Q_{null}(0.95)$. (D) plot of first two principal components of features extracted from the global pooling layer of the computer vision model.}\label{fig:dino-check}
\end{figure}

\begin{figure}[!h]

{\centering \includegraphics[width=1\linewidth]{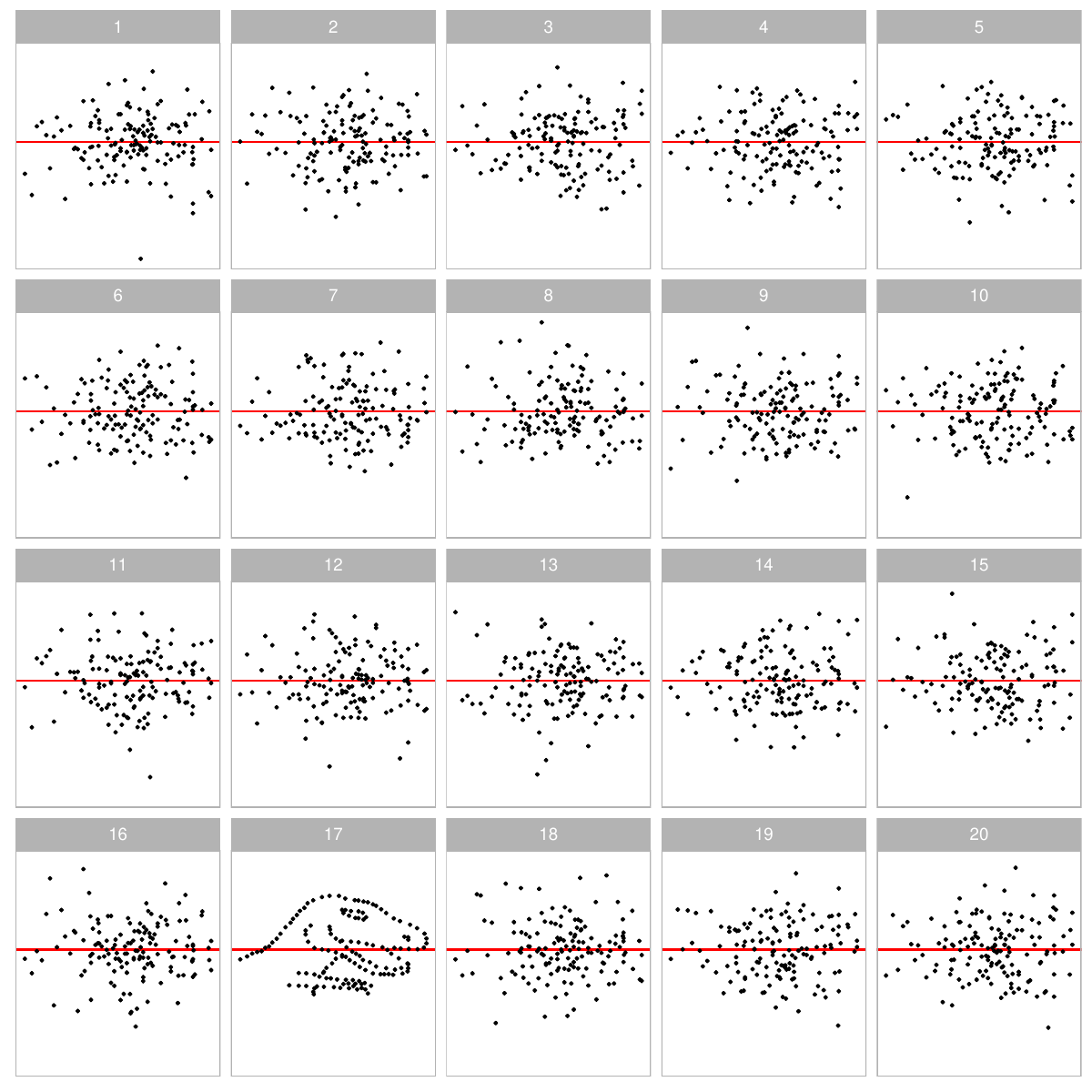} 

}

\caption{A lineup of residual plots for the fitted model on the "dinosaur" dataset. The true residual plot is at position 17. It can be easily identified as the most different plot as the visual pattern is extremely artificial.}\label{fig:dino-lineup}
\end{figure}

The ``dino'' dataset within the \texttt{datasauRus} R package
exemplifies this scenario. With only two columns, x and y, fitting a
regression model to this data yields a residual plot resembling a
``dinosaur'', as displayed in Figure \ref{fig:dino-check}A.
Unsurprisingly, this distinct residual plot stands out in a lineup, as
shown in Figure \ref{fig:dino-lineup}. A visual test conducted by humans
would undoubtedly reject \(H_0\).

According to the residual plot assessment by the computer vision model,
\(\hat{D}\) exceeds \(Q_{null}(0.95)\), warranting a rejection of
\(H_0\). Additionally, most of the bootstrapped fitted models will be
rejected, indicating an misspecified model. However, both the RESET test
and the Breusch-Pagan test yield \(p\)-values greater than 0.3, leading
to a non-rejection of \(H_0\). Only the Shapiro-Wilk test rejects the
normality assumption with a small \(p\)-value.

More importantly, the attention map in Figure \ref{fig:dino-check}B
clearly exhibits a ``dinosaur'' shape, strongly suggesting the
prediction of the distance is based on human perceptible visual
patterns. The computer vision model is also capable of capturing the
contour or the outline of the embedded shape, just like human being
reading residual plots. The PCA in Figure \ref{fig:dino-check}D also
shows that the cluster of bootstrapped plots is at the corner of the
cluster of null plots.

More importantly, the attention map in Figure \ref{fig:dino-check}B
clearly exhibits a ``dinosaur'' shape, strongly suggesting that the
distance prediction is based on human-perceptible visual patterns. The
computer vision model effectively captures the contour or outline of the
embedded shape, similar to how humans interpret residual plots.
Additionally, the PCA in Figure \ref{fig:dino-check}D demonstrates that
the cluster of bootstrapped plots is positioned at the corner of the
cluster of null plots.

In practice, without accessing the residual plot, it would be
challenging to identify the artificial pattern of the residuals.
Moreover, conducting a normality test for a fitted regression model is
not always standard practice among analysts. Even when performed,
violating the normality assumption is sometimes deemed acceptable,
especially considering the application of quasi-maximum likelihood
estimation in linear regression. This example underscores the importance
of evaluating residual plots and highlights how the proposed computer
vision model can facilitate this process.

\section{Limitations and Future Work}\label{limitations-and-future-work}

Despite the computer vision model performing well with general cases
under the synthetic data generation scheme and the three examples used
in this paper, this study has several limitations that could guide
future work.

The proposed distance measure assumes that the true model is a classical
normal linear regression model, which can be restrictive. Although this
paper does not address the relaxation of this assumption, there are
potential methods to evaluate other types of regression models. The most
comprehensive approach would be to define a distance measure for each
different class of regression model and then train the computer vision
model following the methodology described in this paper. To accelerate
training, one could use the convolutional blocks of our trained model as
a feature extractor and perform transfer learning on top of it, as these
blocks effectively capture shapes in residual plots. Another approach
would be to transform the residuals so they are roughly normally
distributed and have constant variance. If only raw residuals are used,
the distance-based statistical testing compares the difference in
distance to a classical normal linear regression model for the true plot
and null plots. This comparison is meaningful only if the difference can
be identified by the distance measure proposed in this paper.

There are other types of residual plots commonly used in diagnostics,
such as residuals vs.~predictor and quantile-quantile plots. In this
study, we focused on the most commonly used residual plot as a starting
point for exploring the new field of automated visual inference.
Similarly, we did not explore other, more sophisticated computer vision
model architectures and specifications for the same reason. While the
performance of the computer vision model is acceptable, there is still
room for improvement to achieve behavior more closely resembling that of
humans interpreting residual plots. This may require external survey
data or human subject experiment data to understand the fundamental
differences between our implementation and human evaluation.

\section{Conclusions}\label{conclusions}

In this paper, we have introduced a distance measure based on
Kullback-Leibler divergence to quantify the disparity between the
residual distribution of a fitted classical normal linear regression
model and the reference residual distribution assumed under correct
model specification. This distance measure effectively captures the
magnitude of model violations in misspecified models. We propose a
computer vision model to estimate this distance, utilizing the residual
plot of the fitted model as input. The resulting estimated distance
serves as the foundation for constructing a single Model Violations
Index (MVI), facilitating the quantification of various model
violations.

Moreover, the estimated distance enables the development of a formal
statistical testing procedure by evaluating a large number of null plots
generated from the fitted model. Additionally, employing bootstrapping
techniques and refitting the regression model allows us to ascertain how
frequently the fitted model is considered misspecified if data were
repeatedly obtained from the same data generating process.

The trained computer vision model demonstrates strong performance on
both the training and test sets, although it exhibits slightly lower
performance on residual plots with non-linearity visual patterns
compared to other types of violations. The statistical tests relying on
the estimated distance predicted by the computer vision model exhibit
lower sensitivity compared to conventional tests but higher sensitivity
compared to visual tests conducted by humans. While the estimated
distance generally mirrors the strength of the visual signal perceived
by humans, there remains scope for further improvement in its
performance.

Several examples are provided to showcase the effectiveness of the
proposed method across different scenarios, emphasizing the similarity
between visual tests and distance-based tests. Overall, both visual
tests and distance-based tests can be viewed as ensemble of tests,
aiming to assess any violations of model assumptions collectively. In
contrast, individual residual diagnostic tests such as the RESET test
and the Breusch-Pagan test only evaluate specific violations of model
assumptions. In practice, selecting an appropriate set of statistical
tests for regression diagnostics can be challenging, particularly given
the necessity of adjusting the significance level for each test.

Our method holds significant value as it helps alleviate a portion of
analysts' workload associated with assessing residual plots. While we
recommend analysts to continue reading residual plots whenever feasible,
as they offer invaluable insights, our approach serves as a valuable
tool for automating the diagnostic process or for supplementary purposes
when needed.

\section*{Acknowledgement}\label{acknowledgement}
\addcontentsline{toc}{section}{Acknowledgement}

These \texttt{R} packages were used for the work: \texttt{tidyverse}
\citep{tidyverse}, \texttt{lmtest} \citep{lmtest}, \texttt{mpoly}
\citep{mpoly}, \texttt{ggmosaic} \citep{ggmosaic}, \texttt{kableExtra}
\citep{kableextra}, \texttt{patchwork} \citep{patchwork},
\texttt{rcartocolor} \citep{rcartocolor}, \texttt{glue} \citep{glue},
\texttt{ggpcp} \citep{ggpcp}, \texttt{here} \citep{here},
\texttt{magick} \citep{magick}, \texttt{yardstick} \citep{yardstick} and
\texttt{reticulate} \citep{reticulate}.

The article was created with R packages \texttt{rticles}
\citep{rticles}, \texttt{knitr} \citep{knitr} and \texttt{rmarkdown}
\citep{rmarkdown}. The project's GitHub repository
(\url{https://github.com/TengMCing/auto_residual_reading_paper})
contains all materials required to reproduce this article.

\bibliographystyle{tfcad}
\bibliography{bibliography.bib}

\begin{thebibliography}{71}
\newcommand{\enquote}[1]{``#1''}
\providecommand{\natexlab}[1]{#1}
\providecommand{\url}[1]{\normalfont{#1}}
\providecommand{\urlprefix}{}

\bibitem[Abadi et~al.(2016)]{abadi2016tensorflow}
Abadi, Mart{\'\i}n, Ashish Agarwal, Paul Barham, Eugene Brevdo, Zhifeng Chen, Craig Citro, Greg~S Corrado, et~al. 2016. ``Tensorflow: Large-scale machine learning on heterogeneous distributed systems.'' \emph{arXiv preprint arXiv:1603.04467} .

\bibitem[Allaire et~al.(2022)]{rticles}
Allaire, JJ, Yihui Xie, Christophe Dervieux, {R Foundation}, Hadley Wickham, {Journal of Statistical Software}, Ramnath Vaidyanathan, et~al. 2022. \emph{rticles: Article formats for R Markdown}. R package version 0.24,  \urlprefix\url{https://CRAN.R-project.org/package=rticles}.

\bibitem[Belsley, Kuh, and Welsch(1980)]{belsley1980regression}
Belsley, David~A, Edwin Kuh, and Roy~E Welsch. 1980. \emph{Regression diagnostics: Identifying influential data and sources of collinearity}. John Wiley \& Sons.

\bibitem[Breusch and Pagan(1979)]{breusch1979simple}
Breusch, Trevor~S, and Adrian~R Pagan. 1979. ``A simple test for heteroscedasticity and random coefficient variation.'' \emph{Econometrica: Journal of the Econometric Society} 1287--1294.

\bibitem[Brunetti et~al.(2018)]{brunetti2018computer}
Brunetti, Antonio, Domenico Buongiorno, Gianpaolo~Francesco Trotta, and Vitoantonio Bevilacqua. 2018. ``Computer vision and deep learning techniques for pedestrian detection and tracking: A survey.'' \emph{Neurocomputing} 300: 17--33.

\bibitem[Buja et~al.(2009)]{buja2009statistical}
Buja, Andreas, Dianne Cook, Heike Hofmann, Michael Lawrence, Eun-Kyung Lee, Deborah~F Swayne, and Hadley Wickham. 2009. ``Statistical inference for exploratory data analysis and model diagnostics.'' \emph{Philosophical Transactions of the Royal Society A: Mathematical, Physical and Engineering Sciences} 367 (1906): 4361--4383.

\bibitem[Chen, Su, and Yang(2020)]{chen2020convolutional}
Chen, Yun, Shijie Su, and Hui Yang. 2020. ``Convolutional neural network analysis of recurrence plots for anomaly detection.'' \emph{International Journal of Bifurcation and Chaos} 30 (01): 2050002.

\bibitem[Chollet et~al.(2015)]{chollet2015keras}
Chollet, Fran\c{c}ois, et~al. 2015. ``Keras.'' \url{https://keras.io}.

\bibitem[Chopra, Hadsell, and LeCun(2005)]{chopra2005learning}
Chopra, Sumit, Raia Hadsell, and Yann LeCun. 2005. ``Learning a similarity metric discriminatively, with application to face verification.'' In \emph{2005 IEEE computer society conference on computer vision and pattern recognition (CVPR'05)}, Vol.~1, 539--546. IEEE.

\bibitem[Chowdhury et~al.(2018)]{chowdhury2018measuring}
Chowdhury, Niladri~Roy, Dianne Cook, Heike Hofmann, and Mahbubul Majumder. 2018. ``Measuring lineup difficulty by matching distance metrics with subject choices in crowd-sourced data.'' \emph{Journal of Computational and Graphical Statistics} 27 (1): 132--145.

\bibitem[Chu et~al.(2019)]{chu2019automatic}
Chu, Hongyang, Xinwei Liao, Peng Dong, Zhiming Chen, Xiaoliang Zhao, and Jiandong Zou. 2019. ``An automatic classification method of well testing plot based on convolutional neural network (CNN).'' \emph{Energies} 12 (15): 2846.

\bibitem[Cook and Weisberg(1982)]{cook1982residuals}
Cook, R~Dennis, and Sanford Weisberg. 1982. \emph{Residuals and influence in regression}. New York: Chapman and Hall.

\bibitem[Davies, Locke, and {D'Agostino McGowan}(2022)]{datasaurus}
Davies, Rhian, Steph Locke, and Lucy {D'Agostino McGowan}. 2022. \emph{datasauRus: Datasets from the Datasaurus Dozen}. R package version 0.1.6,  \urlprefix\url{https://CRAN.R-project.org/package=datasauRus}.

\bibitem[Davison and Hinkley(1997)]{davison1997bootstrap}
Davison, Anthony~Christopher, and David~Victor Hinkley. 1997. \emph{Bootstrap methods and their application}. Cambridge university press.

\bibitem[Efron and Tibshirani(1994)]{efron1994introduction}
Efron, Bradley, and Robert~J Tibshirani. 1994. \emph{An introduction to the bootstrap}. Chapman and Hall/CRC.

\bibitem[Emami and Suciu(2012)]{emami2012facial}
Emami, Shervin, and Valentin~Petrut Suciu. 2012. ``Facial recognition using OpenCV.'' \emph{Journal of Mobile, Embedded and Distributed Systems} 4 (1): 38--43.

\bibitem[Fieberg, Freeman, and Signer(2024)]{fieberg2024using}
Fieberg, John, Smith Freeman, and Johannes Signer. 2024. ``Using lineups to evaluate goodness of fit of animal movement models.'' \emph{Methods in Ecology and Evolution} .

\bibitem[Frisch and Waugh(1933)]{frisch1933partial}
Frisch, Ragnar, and Frederick~V Waugh. 1933. ``Partial time regressions as compared with individual trends.'' \emph{Econometrica: Journal of the Econometric Society} 387--401.

\bibitem[Fukushima and Miyake(1982)]{fukushima1982neocognitron}
Fukushima, Kunihiko, and Sei Miyake. 1982. ``Neocognitron: A new algorithm for pattern recognition tolerant of deformations and shifts in position.'' \emph{Pattern recognition} 15 (6): 455--469.

\bibitem[Gebhardt, Bivand, and Sinclair(2023)]{Albrecht2023interp}
Gebhardt, Albrecht, Roger Bivand, and David Sinclair. 2023. \emph{interp: Interpolation Methods}. R package version 1.1-5,  \urlprefix\url{https://CRAN.R-project.org/package=interp}.

\bibitem[Goodfellow, Bengio, and Courville(2016)]{goodfellow2016deep}
Goodfellow, Ian, Yoshua Bengio, and Aaron Courville. 2016. \emph{Deep learning}. MIT press.

\bibitem[Goscinski et~al.(2014)]{goscinski2014multi}
Goscinski, Wojtek~J, Paul McIntosh, Ulrich Felzmann, Anton Maksimenko, Christopher~J Hall, Timur Gureyev, Darren Thompson, et~al. 2014. ``The multi-modal Australian ScienceS Imaging and Visualization Environment (MASSIVE) high performance computing infrastructure: applications in neuroscience and neuroinformatics research.'' \emph{Frontiers in Neuroinformatics} 8: 30.

\bibitem[Hailesilassie(2019)]{hailesilassie2019financial}
Hailesilassie, Tameru. 2019. ``Financial Market Prediction Using Recurrence Plot and Convolutional Neural Network.''  .

\bibitem[Harrison~Jr and Rubinfeld(1978)]{harrison1978hedonic}
Harrison~Jr, David, and Daniel~L Rubinfeld. 1978. ``Hedonic housing prices and the demand for clean air.'' \emph{Journal of environmental economics and management} 5 (1): 81--102.

\bibitem[Hastie(2017)]{hastie2017generalized}
Hastie, Trevor~J. 2017. ``Generalized additive models.'' In \emph{Statistical models in S}, 249--307. Routledge.

\bibitem[Hatami, Gavet, and Debayle(2018)]{hatami2018classification}
Hatami, Nima, Yann Gavet, and Johan Debayle. 2018. ``Classification of time-series images using deep convolutional neural networks.'' In \emph{Tenth international conference on machine vision (ICMV 2017)}, Vol. 10696, 242--249. SPIE.

\bibitem[He et~al.(2016)]{he2016deep}
He, Kaiming, Xiangyu Zhang, Shaoqing Ren, and Jian Sun. 2016. ``Deep residual learning for image recognition.'' In \emph{Proceedings of the IEEE conference on computer vision and pattern recognition}, 770--778.

\bibitem[Hermite(1864)]{hermite1864nouveau}
Hermite, M. 1864. \emph{Sur un nouveau d{\'e}veloppement en s{\'e}rie des fonctions}. Imprimerie de Gauthier-Villars.

\bibitem[Hester and Bryan(2022)]{glue}
Hester, Jim, and Jennifer Bryan. 2022. \emph{glue: Interpreted String Literals}. R package version 1.6.2,  \urlprefix\url{https://CRAN.R-project.org/package=glue}.

\bibitem[Hofmann, VanderPlas, and Ge(2022)]{ggpcp}
Hofmann, Heike, Susan VanderPlas, and Yawei Ge. 2022. \emph{ggpcp: Parallel Coordinate Plots in the 'ggplot2' Framework}. R package version 0.2.0,  \urlprefix\url{https://CRAN.R-project.org/package=ggpcp}.

\bibitem[Huang et~al.(2017)]{huang2017densely}
Huang, Gao, Zhuang Liu, Laurens Van Der~Maaten, and Kilian~Q Weinberger. 2017. ``Densely connected convolutional networks.'' In \emph{Proceedings of the IEEE conference on computer vision and pattern recognition}, 4700--4708.

\bibitem[Hyndman and Fan(1996)]{hyndman1996sample}
Hyndman, Rob~J, and Yanan Fan. 1996. ``Sample quantiles in statistical packages.'' \emph{The American Statistician} 50 (4): 361--365.

\bibitem[Jeppson, Hofmann, and Cook(2021)]{ggmosaic}
Jeppson, Haley, Heike Hofmann, and Di~Cook. 2021. \emph{ggmosaic: Mosaic plots in the 'ggplot2' framework}. R package version 0.3.3,  \urlprefix\url{https://CRAN.R-project.org/package=ggmosaic}.

\bibitem[Kahle(2013)]{mpoly}
Kahle, David. 2013. ``mpoly: Multivariate Polynomials in {R}.'' \emph{The R Journal} 5 (1): 162--170.

\bibitem[Kingma and Ba(2014)]{kingma2014adam}
Kingma, Diederik~P, and Jimmy Ba. 2014. ``Adam: A method for stochastic optimization.'' \emph{arXiv preprint arXiv:1412.6980} .

\bibitem[Krishnan and Hofmann(2021)]{krishnan2021hierarchical}
Krishnan, Ganesh, and Heike Hofmann. 2021. ``Hierarchical Decision Ensembles-An inferential framework for uncertain Human-AI collaboration in forensic examinations.'' \emph{arXiv preprint arXiv:2111.01131} .

\bibitem[Kuhn, Vaughan, and Hvitfeldt(2024)]{yardstick}
Kuhn, Max, Davis Vaughan, and Emil Hvitfeldt. 2024. \emph{yardstick: Tidy Characterizations of Model Performance}. R package version 1.3.1,  \urlprefix\url{https://CRAN.R-project.org/package=yardstick}.

\bibitem[Kullback and Leibler(1951)]{kullback1951information}
Kullback, Solomon, and Richard~A Leibler. 1951. ``On information and sufficiency.'' \emph{The Annals of Mathematical Statistics} 22 (1): 79--86.

\bibitem[Langsrud(2005)]{langsrud2005rotation}
Langsrud, {\O}yvind. 2005. ``Rotation tests.'' \emph{Statistics and computing} 15: 53--60.

\bibitem[Lee and Chen(2015)]{lee2015image}
Lee, Howard, and Yi-Ping~Phoebe Chen. 2015. ``Image based computer aided diagnosis system for cancer detection.'' \emph{Expert Systems with Applications} 42 (12): 5356--5365.

\bibitem[Li et~al.(2024)]{li2024plot}
Li, Weihao, Dianne Cook, Emi Tanaka, and Susan VanderPlas. 2024. ``A plot is worth a thousand tests: Assessing residual diagnostics with the lineup protocol.'' \emph{Journal of Computational and Graphical Statistics} 1--19.

\bibitem[Loy and Hofmann(2013)]{loy2013diagnostic}
Loy, Adam, and Heike Hofmann. 2013. ``Diagnostic tools for hierarchical linear models.'' \emph{Wiley Interdisciplinary Reviews: Computational Statistics} 5 (1): 48--61.

\bibitem[Loy and Hofmann(2014)]{loy2014hlmdiag}
Loy, Adam, and Heike Hofmann. 2014. ``HLMdiag: A suite of diagnostics for hierarchical linear models in R.'' \emph{Journal of Statistical Software} 56: 1--28.

\bibitem[Loy and Hofmann(2015)]{loy2015you}
Loy, Adam, and Heike Hofmann. 2015. ``Are you normal? The problem of confounded residual structures in hierarchical linear models.'' \emph{Journal of Computational and Graphical Statistics} 24 (4): 1191--1209.

\bibitem[Mason et~al.(2022)]{mason2022cassowaryr}
Mason, Harriet, Stuart Lee, Ursula Laa, and Dianne Cook. 2022. \emph{cassowaryr: Compute Scagnostics on Pairs of Numeric Variables in a Data Set}. R package version 2.0.0,  \urlprefix\url{https://CRAN.R-project.org/package=cassowary}.

\bibitem[Müller(2020)]{here}
Müller, Kirill. 2020. \emph{here: A simpler way to find your files}. R package version 1.0.1,  \urlprefix\url{https://CRAN.R-project.org/package=here}.

\bibitem[Nowosad(2018)]{rcartocolor}
Nowosad, Jakub. 2018. \emph{'CARTOColors' palettes}. R package version 1.0,  \urlprefix\url{https://nowosad.github.io/rcartocolor}.

\bibitem[Ojeda, Solano, and Peramo(2020)]{ojeda2020multivariate}
Ojeda, Sun Arthur~A, Geoffrey~A Solano, and Elmer~C Peramo. 2020. ``Multivariate time series imaging for short-term precipitation forecasting using convolutional neural networks.'' In \emph{2020 International Conference on Artificial Intelligence in Information and Communication (ICAIIC)}, 33--38. IEEE.

\bibitem[O'Malley et~al.(2019)]{omalley2019kerastuner}
O'Malley, Tom, Elie Bursztein, James Long, Fran\c{c}ois Chollet, Haifeng Jin, Luca Invernizzi, et~al. 2019. ``Keras {Tuner}.'' \url{https://github.com/keras-team/keras-tuner}.

\bibitem[Ooms(2023)]{magick}
Ooms, Jeroen. 2023. \emph{magick: Advanced Graphics and Image-Processing in R}. R package version 2.7.4,  \urlprefix\url{https://CRAN.R-project.org/package=magick}.

\bibitem[Pedersen(2022)]{patchwork}
Pedersen, Thomas~Lin. 2022. \emph{patchwork: The composer of plots}. R package version 1.1.2,  \urlprefix\url{https://CRAN.R-project.org/package=patchwork}.

\bibitem[Ramsey(1969)]{ramsey1969tests}
Ramsey, James~Bernard. 1969. ``Tests for specification errors in classical linear least-squares regression analysis.'' \emph{Journal of the Royal Statistical Society: Series B (Methodological)} 31 (2): 350--371.

\bibitem[Rawat and Wang(2017)]{rawat2017deep}
Rawat, Waseem, and Zenghui Wang. 2017. ``Deep convolutional neural networks for image classification: A comprehensive review.'' \emph{Neural computation} 29 (9): 2352--2449.

\bibitem[Series(2011)]{series2011studio}
Series, BT. 2011. ``Studio encoding parameters of digital television for standard 4: 3 and wide-screen 16: 9 aspect ratios.'' \emph{International Telecommunication Union, Radiocommunication Sector} .

\bibitem[Shapiro and Wilk(1965)]{shapiro1965analysis}
Shapiro, Samuel~Sanford, and Martin~B Wilk. 1965. ``An analysis of variance test for normality (complete samples).'' \emph{Biometrika} 52 (3/4): 591--611.

\bibitem[Silverman(2018)]{silverman2018density}
Silverman, Bernard~W. 2018. \emph{Density estimation for statistics and data analysis}. Routledge.

\bibitem[Simonyan and Zisserman(2014)]{simonyan2014very}
Simonyan, Karen, and Andrew Zisserman. 2014. ``Very deep convolutional networks for large-scale image recognition.'' \emph{arXiv preprint arXiv:1409.1556} .

\bibitem[Singh et~al.(2017)]{singh2017deep}
Singh, Karamjit, Garima Gupta, Lovekesh Vig, Gautam Shroff, and Puneet Agarwal. 2017. ``Deep convolutional neural networks for pairwise causality.'' \emph{arXiv preprint arXiv:1701.00597} .

\bibitem[Srivastava et~al.(2014)]{srivastava2014dropout}
Srivastava, Nitish, Geoffrey Hinton, Alex Krizhevsky, Ilya Sutskever, and Ruslan Salakhutdinov. 2014. ``Dropout: a simple way to prevent neural networks from overfitting.'' \emph{The journal of machine learning research} 15 (1): 1929--1958.

\bibitem[Tukey and Tukey(1985)]{tukey1985computer}
Tukey, John~W, and Paul~A Tukey. 1985. ``Computer graphics and exploratory data analysis: An introduction.'' In \emph{Proceedings of the sixth annual conference and exposition: computer graphics}, Vol.~85, 773--785.

\bibitem[Ushey, Allaire, and Tang(2024)]{reticulate}
Ushey, Kevin, JJ~Allaire, and Yuan Tang. 2024. \emph{reticulate: Interface to 'Python'}. R package version 1.35.0,  \urlprefix\url{https://CRAN.R-project.org/package=reticulate}.

\bibitem[Vo and Hays(2016)]{vo2016localizing}
Vo, Nam~N, and James Hays. 2016. ``Localizing and orienting street views using overhead imagery.'' In \emph{Computer Vision--ECCV 2016: 14th European Conference, Amsterdam, The Netherlands, October 11--14, 2016, Proceedings, Part I 14}, 494--509. Springer.

\bibitem[Wang et~al.(2004)]{wang2004image}
Wang, Zhou, Alan~C Bovik, Hamid~R Sheikh, and Eero~P Simoncelli. 2004. ``Image quality assessment: from error visibility to structural similarity.'' \emph{IEEE transactions on image processing} 13 (4): 600--612.

\bibitem[Wickham et~al.(2019)]{tidyverse}
Wickham, Hadley, Mara Averick, Jennifer Bryan, Winston Chang, Lucy~D'Agostino McGowan, Romain François, Garrett Grolemund, et~al. 2019. ``Welcome to the {tidyverse}.'' \emph{Journal of Open Source Software} 4 (43): 1686.

\bibitem[Widen et~al.(2016)]{widen2016graphical}
Widen, Holly~M, James~B Elsner, Stephanie Pau, and Christopher~K Uejio. 2016. ``Graphical inference in geographical research.'' \emph{Geographical Analysis} 48 (2): 115--131.

\bibitem[Wilkinson, Anand, and Grossman(2005)]{wilkinson2005graph}
Wilkinson, Leland, Anushka Anand, and Robert Grossman. 2005. ``Graph-theoretic scagnostics.'' In \emph{Information Visualization, IEEE Symposium on}, 21--21. IEEE Computer Society.

\bibitem[Xie(2014)]{knitr}
Xie, Yihui. 2014. ``knitr: A comprehensive tool for reproducible research in {R}.'' In \emph{Implementing reproducible computational research},  edited by Victoria Stodden, Friedrich Leisch, and Roger~D. Peng. Chapman and Hall/CRC. ISBN 978-1466561595,  \urlprefix\url{http://www.crcpress.com/product/isbn/9781466561595}.

\bibitem[Xie, Dervieux, and Riederer(2020)]{rmarkdown}
Xie, Yihui, Christophe Dervieux, and Emily Riederer. 2020. \emph{R Markdown cookbook}. Boca Raton, Florida: Chapman and Hall/CRC. ISBN 9780367563837,  \urlprefix\url{https://bookdown.org/yihui/rmarkdown-cookbook}.

\bibitem[Zeileis and Hothorn(2002)]{lmtest}
Zeileis, Achim, and Torsten Hothorn. 2002. ``Diagnostic checking in regression relationships.'' \emph{R News} 2 (3): 7--10.

\bibitem[Zhang et~al.(2020)]{zhang2020encoding}
Zhang, Ye, Yi~Hou, Shilin Zhou, and Kewei Ouyang. 2020. ``Encoding time series as multi-scale signed recurrence plots for classification using fully convolutional networks.'' \emph{Sensors} 20 (14): 3818.

\bibitem[Zhu(2021)]{kableextra}
Zhu, Hao. 2021. \emph{kableExtra: Construct complex table with kable and pipe syntax}. R package version 1.3.4,  \urlprefix\url{https://CRAN.R-project.org/package=kableExtra}.

\end{thebibliography}

\end{document}